\documentclass{deepmind}

\title{Acme: A Research Framework for Distributed Reinforcement Learning}

\author{
{\Authfont 
Matthew W.\ Hoffman\textsuperscript{*\textdagger},
Bobak Shahriari\textsuperscript{*\textdagger},
John Aslanides\textsuperscript{\textdagger},
Gabriel Barth-Maron\textsuperscript{\textdagger}}\protect\\
{\Affilfont DeepMind}
\protect\\[1em]
{\Authfont
Nikola Momchev\textsuperscript{*},
Danila Sinopalnikov\textsuperscript{*},
Piotr Sta\'nczyk\textsuperscript{*},
Sabela Ramos, Anton Raichuk, Damien Vincent}
\protect\\
{\Affilfont Google Research, Brain Team}
\protect\\[1em]
{\Authfont 
L\'eonard Hussenot\textsuperscript{*},
Robert Dadashi\textsuperscript{*},
Gabriel Dulac-Arnold, Manu Orsini, Alexis Jacq, Johan Ferret, Nino Vieillard,
Seyed Kamyar Seyed Ghasemipour, Sertan Girgin, Olivier Pietquin}
\protect\\
{\Affilfont Google Research, Brain Team}
\protect\\[1em]
{\Authfont Feryal Behbahani, Tamara Norman, Abbas Abdolmaleki, Albin Cassirer,
Fan Yang, Kate Baumli, Sarah Henderson, Abe Friesen,
Ruba Haroun\textsuperscript{\textdaggerdbl},
Alex Novikov, Sergio G\'omez Colmenarejo, Serkan Cabi, Caglar Gulcehre,
Tom Le Paine, Srivatsan Srinivasan, Andrew Cowie,
Ziyu Wang\textsuperscript{\textdaggerdbl},
Bilal Piot, Nando de Freitas}
\protect\\
{\Affilfont DeepMind, \textsuperscript{\textdaggerdbl}Work done while at DeepMind}
\protect\\[1em]
{\Affilfont
\textsuperscript{*}Core Contributor\protect\\
\textsuperscript{\textdagger}Original Author
}}

\usepackage[utf8]{inputenc}
\usepackage{placeins}
\usepackage{natbib}
\usepackage{graphicx}
\usepackage{listings}
\usepackage{tcolorbox}
\usepackage{enumitem}
\usepackage{subfigure}
\usepackage{algorithm,algorithmic}
\usepackage{sidecap}
\sidecaptionvpos{figure}{c}

\floatname{algorithm}{Listing}
\graphicspath{{figures/}}

\setlist{topsep=0pt, partopsep=0pt, parsep=0pt, itemsep=0.5em}

\DeclareMathOperator*{\argmax}{arg\,max}

\newcommand{\E}{\mathbb{E}}

\newtcbox{\mybox}[1][red]{on line,
arc=3pt,colback=#1!10!white,colframe=white,colback=#1!70!,
before upper={\rule[-3pt]{0pt}{10pt}},boxrule=1pt,
boxsep=0pt,left=3pt,right=3pt,top=2pt,bottom=1pt}

\newcommand{\mytag}[2]{\mybox[#1]{\texttt{\scriptsize #2}}}

\newcommand{\tagVnetwork}{\mytag{Lavender}{V-Network}}
\newcommand{\tagQnetwork}{\mytag{Lavender}{Q-Network}}
\newcommand{\tagPnetwork}{\mytag{Lavender}{Policy-Network}}
\newcommand{\tagoffpolicy}{\mytag{YellowGreen}{Off-Policy}}
\newcommand{\tagonpolicy}{\mytag{YellowGreen}{On-Policy}}
\newcommand{\tagoffline}{\mytag{YellowGreen}{Offline}}
\newcommand{\tagbootstrapping}{\mytag{Bittersweet}{Bootstrapping}}
\newcommand{\tagmc}{\mytag{Bittersweet}{MC}}
\newcommand{\tagdiscrete}{\mytag{Goldenrod}{Discrete Actions}}
\newcommand{\tagcontinuous}{\mytag{Goldenrod}{Continuous Actions}}
\newcommand{\tagdiscretecontinuous}{\mytag{Goldenrod}{Discrete \& Continuous Actions}}

\begin{abstract}
Deep reinforcement learning (RL) has led to many recent and groundbreaking
advances. However, these advances have often come at the cost of both
increased scale in the underlying architectures being trained as well as
increased complexity of the RL algorithms used to train them. These increases
have in turn made it more difficult for researchers to rapidly prototype new
ideas or reproduce published RL algorithms. To address these concerns this
work describes Acme, a framework for constructing novel RL algorithms that is
specifically designed to enable agents that are built using simple, modular
components that can be used at various scales of execution. While the primary
goal of Acme is to provide a framework for algorithm development, a secondary
goal is to provide simple reference implementations of important or
state-of-the-art algorithms. These implementations serve both as a validation
of our design decisions as well as an important contribution to
reproducibility in RL research. In this work we describe the major design
decisions made within Acme and give further details as to how its components
can be used to implement various algorithms. Our experiments provide baselines
for a number of common and state-of-the-art algorithms as well as showing how
these algorithms can be scaled up for much larger and more complex
environments. This highlights one of the primary advantages of Acme, namely
that it can be used to implement large, distributed RL algorithms that can run
at massive scales while still maintaining the inherent readability of that
implementation.

\vspace{1em}
This work presents a second version of the paper which coincides with an
increase in modularity, additional emphasis on offline, imitation and learning
from demonstrations algorithms, as well as various new agents implemented as
part of Acme.
\end{abstract}

\begin{document}
\maketitle
\tableofcontents

\section{Introduction}

Reinforcement learning (RL) provides an elegant formalization with which to
study and design intelligent agents \citep{russell2016rationality}. At its
broadest level we can think of an agent as an object which is able to interact
with its environment, to make observations, and to use these observations in
order to improve its own performance. This formulation and the algorithms
developed to train these agents have a long history that goes back at least to
the 1950s~\citep{bellman1957mdp,bellman1966dynamic}, see
\citet[][Section~1.7]{sutton2018reinforcement} for an excellent and concise
history. Modern \emph{deep RL} combines advances in reinforcement learning
with improved neural network architectures and increased computational
resources. These advances have in turn led to dramatic increases in agent
capabilities. This modern era was perhaps ushered in by huge leaps forward in
game-playing agents \citep{mnih2015atari, silver2016alphago}, a trend that
continues to push the boundaries of RL research to this day
\citep{openai2019five, vinyals2019starcraft}. More recent works have begun to
move toward real-world domains as varied as robotics
\citep{openai2018learning}, stratospheric balloon
navigation~\citep{bellemare2020autonomous}, nuclear fusion
\citep{degrave2022magnetic}, and more. The pursuit of state-of-the-art
performance on such large and varied domains has in turn contributed to a huge
uptick in both the complexity and scale of agents developed by the research
community.

One characteristic that has contributed to the increasing complexity in RL
research has been an integrationist approach to agent design in which modern
agents often involve a combination of an ever-increasing number of smaller
algorithmic components. This has led to a state in which state-of-the-art
agents incorporate numerous independent components. Examples include
modifications to the reward signals to include intrinsic rewards
\citep{bellemare2016unifying,badia2020never} or auxiliary tasks
\citep{jaderberg2017auxillary,jaques2019social}, as well as specialized neural
network architectures \citep{wang2015dueling}, and model ensembles
\citep{osband2018rpf}. Core techniques have been rethought including replay
\citep{schaul2015prioritized}, the form of value function backups
\citep{bellemare2017distributional}, or the use of search within policy
improvement \citep{silver2018alphazero}. Various forms of variance reduction
\citep{wang2016sample, schulman2017proximal, espeholt2018impala}, hierarchical
learning \citep{kulkarni2016hierarchical, vezhnevets2017fun}, or meta-learning
\citep{alshedivat2017continuous,finn2017model,xu2018meta} have
also been proposed.

The performance of modern reinforcement learning algorithms, like much of
machine learning, has also benefited greatly from increases in scale. This
question of scale can be considered along two related dimensions: the capacity
of the function approximator and the amount of training data presented to the
algorithm. The first is an active area of research due to the increased
importance of model parallelism and the latter is of particular importance in
the context of RL as the agent typically has to produce its own data by
interacting with the environment. This motivates distributing data gathering
across multiple parallel instances of an environment in order to generate data
as fast as possible. This has led to the widespread use of increasingly
large-scale distributed systems in RL agent training \citep{mnih2016a3c,
horgan2018apex, espeholt2018impala, kapturowski2019r2d2, espeholt2019seed}.
This approach introduces engineering and algorithmic challenges, and relies on
significant amounts of infrastructure which can impede the reproducibility of
research. This is also the case with the increasing size of models we have
seen in recent years. It also motivates agent designs that may represent
dramatic departures from canonical abstractions laid out in the reinforcement
learning literature \citep{sutton2018reinforcement}. This means that scaling a
simple prototype to a full distributed system may require a complete
re-implementation of the agent.

To combat the growing complexity inherent in modern RL implementations, we
developed \textbf{Acme}, a framework for expressing and training
RL agents which attempts to address both the issues of complexity and scale
within a unified framework. The goal of Acme is to allow for fast
iteration of research ideas and scalable implementation of state-of-the-art
agents. Acme does this by providing tools and components for constructing
agents at various levels of abstraction, from basic modules and functions
(e.g. networks, losses, policies), to workers which update the state of the
learning process (actors, learners, replay buffers), and finally complete
agents with the experimental apparatus necessary for robust measurement and
evaluation, such as  training loops, logging, and checkpointing. The agents
written in the framework are state-of-the-art implementations that promote the
use of common tools and components. Our modular design of Acme's agents makes
them easy to combine or scale to large distributed systems, all while
maintaining clear and straightforward abstractions. It simultaneously supports
training in a simpler synchronous setting which can be both easier to
understand and to debug.

In Section \ref{sec:modern-rl} we give a brief overview of modern
reinforcement learning and discuss various software frameworks used to tackle
such problems. Section~\ref{sec:acme} goes on to introduce the key structural
contributions of our approach to designing RL agents. In
Section~\ref{sec:agents} we build upon this design to show how this framework
is used to implement a number of modern agent implementations. Finally, in
Section~\ref{sec:experiments} we experiment with these agents and demonstrate
that they can be used to obtain state-of-the-art performance across a variety
of domains.

Note that this work also represents a second version of the paper. In the
interim Acme has seen an increase in its modularity due to the design of
\emph{agent builders} (Section~\ref{sec:builders}). We have also put
additional emphasis on offline and imitation algorithms including many
algorithms that make use of the builder design to compose agents. The number
of core algorithms implemented as part of Acme has also increased. Finally,
although it is not the focus of this paper much of the algorithms implemented in
Acme have increasingly transitioned to JAX~\citep{jax2018github} to implement
the network architecture of policies, critics, etc. While much of the base
interfaces are still platform agnostic maintaining one-to-one correspondence
between e.g.\ TensorFlow~\citep{tensorflow2015-whitepaper} and JAX has been
de-emphasized due to its corresponding support burden.

\section{Modern Reinforcement Learning}
\label{sec:modern-rl}

\begin{figure}
\centering
\includegraphics[trim=0cm 9.0cm 16.8cm 0cm, clip]{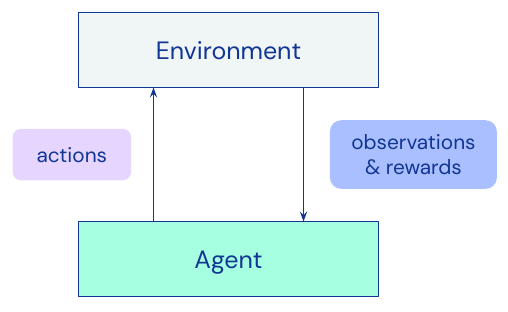}
\caption{A simple, high-level illustration of an agent interacting with its
environment. Here we illustrate the flow of information between the agent and
its environment wherein the environment consumes actions produced by the agent
and returns rewards and subsequent observations.}
\label{fig:environment-loop}
\end{figure}

The standard setting for reinforcement learning involves a learning
\emph{agent}---an entity that perceives and acts---interacting with an
\emph{environment} in discrete time. At a high level this setting can be
thought of as an interactive loop in which the agent produces actions based on
its observations of the environment and the environment consumes those actions
before returning a reward and subsequent observation. See
Figure~\ref{fig:environment-loop} for a brief illustration of this
interaction. From a mathematical perspective, the interaction between an agent
and its environment can generally be characterized as either a Markov Decision
Process (MDP) \citep{puterman1994markov} or a Partially Observable MDP (POMDP)
\citep{kaelbling1998planning}. In both of these formulations the environment
is assumed to be determined at any time $t$ by some \emph{state} $x_t$. In the
MDP setting agents directly observe this state and make decisions based on its
value. The POMDP setting, however, generalizes this assumption such that the
agent is only able to perceive \emph{observations} $o_t$ which are functionally
(or stochastically) generated based on the hidden state. In other words an MDP
can be written more generally as a POMDP where $o_t=x_t$. For a simple example
illustrating this distinction, consider the widely used Atari set of
benchmarks~\citep{bellemare2013ale}, for which $x_t$ corresponds to the
internal state of a given game and $o_t$ corresponds to the pixel observations
rendered from this state. In general we will assume the more general POMDP
setting for all environments, however we leave the question of how to address
partial observability up to particular agent implementations. This assumption
does not fundamentally alter the agent/environment interface illustrated above
and in Section~\ref{sec:environment} we describe this interface and Acme's
implementation of it in more detail.

The mechanism by which an agent interacts with the environment is primarily
determined by its \emph{policy}, i.e.\ a function which maps its experienced
history of observations to an \emph{action} $a_t$. While the policy can be
represented as an arbitrary function, in this work we will assume for
expository reasons that it is a deep neural network of some form. Under this
assumption, the simplest instance of an agent's mapping might be that of a
feed-forward neural network evaluated on the most recent observation. However,
in order to optimally act in more general settings (e.g.\ partially observable
environments), an agent might be implemented using a recurrent network. Here
the recurrent state of the network can act as a proxy for the environment's
internal (hidden) state. Section~\ref{sec:actors} describes in more detail the
way in which policies are represented in Acme and how these are used to
generate an action. As shown in Figure~\ref{fig:environment-loop}, given an
action selected by the agent the environment updates its internal state and
emits rewards and observations; Section~\ref{sec:replay} describes how this
data can then be collected and used to update the agent's behavior.

The typical objective of any RL agent is to optimize its policy $\pi$
so as to maximize some aggregate measure of its expected future rewards. If we
let $R_t$ denote the random variable associated with rewards at time $t$
(i.e.\ the reward yielded by the environment after following the policy $\pi$),
we can then write the \emph{discounted future returns} of this policy as
\begin{equation}
    G_t = R_t + \gamma R_{t+1} + \gamma^2 R_{t+2} + \cdots
    \label{eq:return}
\end{equation}
Here the discount factor $\gamma\in[0,1)$ is simply a term that prioritizes
near-term rewards over longer-term rewards. Note that this choice of
aggregation---i.e.\ summed, discounted rewards---is one particular, albeit
common choice. Most of the agents that we focus on in this work target this
infinite-horizon, discounted reward formulation, however agents in Acme are
capable of optimizing other such measures. This choice is typically made by
the \emph{learner} component of an algorithm that we introduce in
Section~\ref{sec:learners} and we discuss particular implementations of it in
Section~\ref{sec:agents}. Given this measure of performance we can define an
agent's optimal policy as $\pi^*=\argmax_{\pi}\E_\pi[G_t]$, i.e.\ the policy
which leads to the best return in expectation. Here we have explicitly noted
that the expectation is dependent on the policy $\pi$, however we have left
implicit all the other distributions involved, i.e.\ in practice this
expectation should be evaluated with respect to any stochasticity in the
environment as well. The question of how best to estimate and/or directly
optimize this expectation is predominately what leads to different mechanisms
for updating an agent's policy.

Finally, while the description of reinforcement learning given in this section
provides a high level overview, we can also divide this space into a number of
alternative formulations that are also supported by Acme. In what remains of
this section we will highlight a number of these settings.

\paragraph{Online Reinforcement Learning}
The online setting largely corresponds to the description given above, and
could be described more simply as the ``standard'' setting for reinforcement
learning \citep{sutton2018reinforcement}. Under this problem formulation an
agent interacts with its environment and maximizes the accumulated reward
collected through trial-and-error. This formulation emphasizes an agent's
ability to learn by actively exploring its environment and can be contrasted
against settings in which an agent makes more passive observations of the
environment. Additionally, Online RL is often further split into on-policy or
off-policy algorithms, a distinction which refers to whether data is generated
by the same policy that is being optimized (on-policy) or a separate behavior
policy. Finally, although it is not a strict one-to-one correspondence many
off-policy algorithms also make use of experience replay, i.e.\ experience is
stored as it is generated so that updates to the policy can query the entire
history of generated data rather than the most recently generated samples.
Off-policy algorithms also typically rely on what is known as
\emph{bootstrapping} wherein an estimates of the value of a given policy is
used in a dynamic-programming-like fashion to improve its own estimate (i.e.\
the estimates bootstrap upon themselves).

\paragraph{Offline Reinforcement Learning}
The setting of offline RL (also referred to as batch RL)
\citep{ernst2005tree,riedmiller2005neural,lange2012batch,levine2020offline},
is used when the agent cannot interact with its environment, and instead
learns a policy from an existing collection of experience. This setting is
particularly relevant if interactions with the environment are costly or
dangerous and appeals to a natural motivation to facilitate learning by
leveraging historical interactions. However, as policies learned in this
setting do not receive feedback from the environment, such algorithms may
wrongfully associate high returns with actions that were not taken in the
collection of existing experience. Thus, methods in this offline setting
frequently incorporate regularization that incentivizes the the learned policy
to avoid \emph{unknown} or \emph{uncertain} state-action regions. This can
further be contrasted with online methods where uncertainty is often
encouraged as a mechanism to fully explore the state space (often known as
\emph{optimism in the face of uncertainty}).

\paragraph{Imitation Learning}
For various tasks, the environment might not come with a reward which is
well-defined. For instance, suppose that we have a game simulator, and that
the goal is to design a policy that "plays like a beginner". The specification
of a reward function that would explain the behavior of a beginner player is
tedious. In the Imitation Learning (IL) setting \citep{pomerleau1991efficient,
bagnell2007boosting, ross2010efficient}, the environment does not come with a
reward, but instead with a number of demonstrations which are trajectories of
an agent interacting with the environment of interest. The objective is either
to learn a policy whose behavior match the one in the provided demonstrations,
or to infer the reward that best explains the behavior in the provided
demonstrations \citep[see also][]{ng2000algorithms, russell1998learning,
ziebart2008maximum, abbeel2004apprenticeship}.

\paragraph{Learning from Demonstrations}
In the Learning from Demonstrations (LfD) setting~\citep{schaal1997learning,
hester2018deep, vecerik2017leveraging, pmlr-v162-dadashi22a}, similarly to the
standard online RL setting, the goal is to learn a policy through trial and
error by interacting with an environment, which has a well-defined reward to
optimize. In addition, similarly to the IL setting, the learning agent has
access to examples of trajectories that come from an "expert" agent. This is
relevant in domains where the reward function is easy to define but is
challenging to optimize. For example, for tasks with sparse reward signals
(e.g., a manipulation task that yields a positive reward only if the task is
completed) the use of demonstrations facilitates the exploration problem.

\section{Acme}
\label{sec:acme}

At its core Acme is a framework designed to enable readable and efficient
implementations of reinforcement learning algorithms that target the
development of novel RL agents and their applications. One of the hallmarks of
modern reinforcement learning is the scale at which such algorithms are run.
As a result, algorithms implemented with Acme usually culminate in a
\emph{distributed agent} which involves many separate (parallel) acting,
learning, as well as diagnostic and helper processes. However, one of the key
design decisions of Acme is to re-use the same components between simple,
single-process implementations and large, distributed agents. This gives
researchers and algorithm developers the ability to implement RL algorithms
once, regardless of how they will later be executed.

\begin{figure}[b!]
\centering
\includegraphics[width=0.7\textwidth, trim=0cm 7.5cm 12.5cm 0cm, clip]{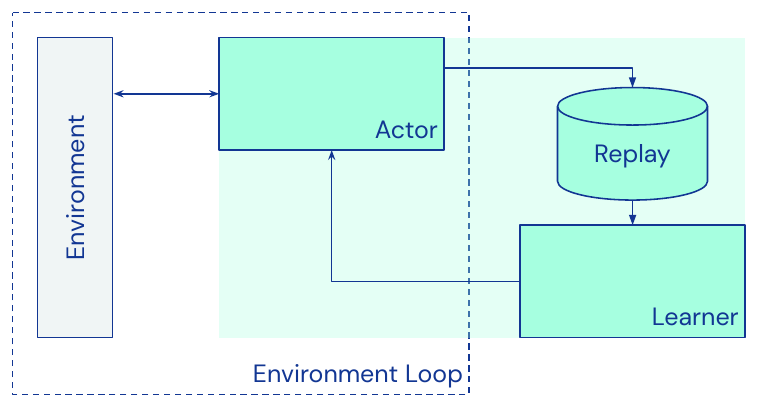}
\caption{An expanded view of the agent-environment interaction shown in
Figure~\ref{fig:environment-loop}. This explicitly displays the environment
loop which controls the interface between the agent and its environment; the
communication between these components is now represented by a single,
bidirectional link. We have also expanded the agent into into its three major
components which include an actor which generates actions and receives
observations, a replay system which stores data produced by the actor, and a
learner which consumes this data to produce parameter updates.}
\label{fig:environment-loop-expanded}
\end{figure}

To keep things simple, we will begin by describing those components that
should be readily familiar to anyone well-versed in the RL literature
\citep[e.g.][]{sutton2018reinforcement} as illustrated in
Figure~\ref{fig:environment-loop-expanded}. First and foremost among these
components is the environment with which the agent interacts, described in
Section~\ref{sec:environment}. The agent itself is then broken down into three
primary components: an \emph{actor} which interacts with the environment to
generate data, a \emph{replay} system which stores that data, and a
\emph{learner} which updates the agent's behavior based on data sampled from
replay; these elements are described in
Sections~\ref{sec:actors}--\ref{sec:learners} respectively. Next, we will show
how these components can be combined to train and evaluate the performance of
a given agent. This includes \emph{builders} which defines how to construct
all the components of an agent (Section~\ref{sec:builders}). We will also
describe experiment runners for both local and distributed agents
(Sections~\ref{sec:running}--\ref{sec:distributed}) along with their shared
configuration. We will then describe how these agents can be reused for
offline settings with limited (or even zero) modification.

Note, in what follows we will give a high-level overview of the core
interfaces and building blocks of Acme. We are, however, very cognizant of the
fact that Acme is a living framework and some of this design (particularly
with respect to function inputs) may subtly shift over time. While the
high-level components, methods, and their uses should be relatively stable,
for up-to-date information on the precise APIs please refer to our
documentation (\url{https://dm-acme.readthedocs.io/}).

\subsection{Environments and environment loops}
\label{sec:environment}

The \emph{environment} with which an agent interacts is a core concept within
reinforcement learning. In Acme we assume an environment which maintains its
own state and that follows an interface defined by the \texttt{dm\_env}
package~\citep{muldal2019dm_env}. Ultimately, an environment is represented by
an object which implements a \texttt{step} function that takes an action $a_t$
and produces a tuple $(r_t, d_t, o_{t+1}, e_{t+1})$ consisting of:
\begin{itemize}
\item the reward $r_t$, provided by the environment as it transitions to its
new state;
\item an environmental discount $0\leq d_t\leq 1$ that allows for
environment-specific discounting of future rewards---this acts like the
discount factor $\gamma$ but is provided by the environment; this term is most
often used to distinguish between episode termination and truncation (see
below);
\item a new observation $o_{t+1}$ that corresponds to any observable
quantities of the new state provided by the environment;
\item and an end-of-episode indicator $e_{t+1}$. Environments here are assumed
to be episodic, and the end-of-episode indicator is a Boolean value which is
true when the step is the final step of an episode, i.e.\ the lifetime of a
single agent instance. This signal is often used to reset any internal agent
state in order to begin a new episode.
\end{itemize}
Note that at the end of an episode (i.e.\ when $e_{t+1}$ is True) we can have
$d_t=0$, which signals the \emph{termination} of the episode and therefore no
more reward(s) can be obtained. This is in contrast to the case where $d_t>0$,
which signals the \emph{truncation} of an episode---i.e.\ additional rewards
could be obtained by continuing to interact with the environment but the
environment is signaling that it will end the episode early and sample a new
one. The distinction between these two cases is particularly relevant when
learning a bootstrapped value function (which we will define more carefully in
Section~\ref{sec:agents}). In particular, bootstrapping \emph{should} be used
when an episode is truncated but not when it is terminated.

The interaction between an agent and its environment is mediated through use
of an \emph{environment loop}. This provides a simple entry point which, given
actor and environment instances can be repeatedly called to interact with the
environment for a given number of transitions or whole episodes. Instances of
the actor interface, which will be introduced in the following section, can be
roughly thought of as a wrapper around a raw policy which maintains the
policy's state. Given these two components,
Listing~\ref{lst:environment-loop-pseudocode} shows both pseudocode and a
corresponding, albeit simplified, implementation of the environment loop. In
particular this highlights the interleaved interaction between the actor and
environment, where the environment generates observations, the actor consumes
those observations to produce actions, and the environment loop maintains the
state of this interaction. A full implementation of this loop largely
corresponds with the given pseudocode, albeit here the pseudocode eliminates
logging, checkpointing, and other bookkeeping operations.

\begin{algorithm}
\centering
\begin{minipage}[t]{0.34\textwidth}
\begin{algorithmic}[1]
    \WHILE{True}
    \STATE{$t\gets0$; $e_t\gets\mathrm{False}$}
    \STATE{Reset environment to get $o_0$}
    \STATE{Observe $o_0$}

    \vspace{\baselineskip}
    \WHILE{not $e_t$}

        \STATE{$a_t\gets\pi(o_t)$}
        \STATE{Take action $a_t$; obtain step\\
            \hskip\algorithmicindent 
            $\eta_t = (r_t, d_t, o_{t+1}, e_{t+1})$}
            
        \vspace{1\baselineskip}
        \STATE{Observe $(a_t, \eta_t)$}
        \STATE{Update the policy $\pi$}
        \STATE{$t\gets t+1$}
    \ENDWHILE
    \ENDWHILE
\end{algorithmic}
\end{minipage}
\begin{minipage}[t]{0.63\textwidth}
\vspace{0em}
\begin{verbatim}
while True:
  # Make an initial observation.
  step = environment.reset()
  actor.observe_first(step)
  
  while not step.last():
    # Evaluate the policy and step the environment.
    action = actor.select_action(step.observation)
    step = environment.step(action)

    # Make an observation and update the actor.
    actor.observe(action, next_timestep=step)
    actor.update()
\end{verbatim}
\end{minipage}
\caption{Pseudocode for a simplified environment loop. Also shown for
comparison is an implementation of this loop using Acme components. The given
code is broadly equivalent to the actual \texttt{EnvironmentLoop} class and is
only lacking in additional bookkeeping operations. \vspace{0.5em}}
\label{lst:environment-loop-pseudocode}
\end{algorithm}

\subsection{Actors}
\label{sec:actors}

In Acme, the agent's interaction with the environment is controlled by an
environment loop, but the process of generating actions is performed by an
\emph{actor}. This component consumes the environment outputs---the rewards,
observations, etc.---and produces actions that are passed to the environment.
While it might be possible to combine this step into a single function, in
Acme we explicitly split this into four methods:
\begin{itemize}
\item \texttt{select\_action} consumes the observation $o_t$ and returns the action
$a_t$ selected by the agent; any internal state for the agent should be
computed and stored by this method.

\item \texttt{observe} consumes the agent's selected action $a_t$ and the
timestep tuple $(r_t, d_t, o_{t+1}, e_{t+1})$ produced by the environment;
this data can then be stored or otherwise used for learning.

\item \texttt{observe\_first} plays the same role as the \texttt{observe} method,
but should be called on the first step of an episode when only the observation
$o_0$ has been produced.

\item \texttt{update} allows the actor to perform any internal parameter
updates. In distributed agents this usually corresponds to requesting the
latest policy parameters from the learner.
\end{itemize} 
By splitting these methods up we can clearly delineate the responsibility and
inputs for each agent function. While it is possible to implement all of an
agent's logic behind the \texttt{Actor} interface, we generally prefer to
limit the use of this interface to only those functions necessary for
generating actions and storing the experience data generated by this
interaction. By separating the responsibilities for data generation and
training (Section~\ref{sec:learners}) the agent can be distributed and
parallelized in a much simpler fashion as we will demonstrate in
Section~\ref{sec:distributed}. With this in mind, the three primary methods
defined by the actor interface can be seen as evaluating the policy, storing
observed data, and retrieving updated parameters from an external variable
source.

While the \texttt{select\_action} of an actor is in charge in charge of
evaluating the agent's policy in order to select the next action, the policy
is usually being learned (represented by its parameters) and we must be able
to handle changes to these parameters when the \texttt{update} method is
called. In order to seamlessly support both single-process and distributed
execution, the processing of updating parameters on each \texttt{Actor} is
performed by periodically polling some variable source for the most recent
parameters. When an \texttt{Actor} is instantiated it is given an object
implementing the \texttt{VariableSource} interface. This interface includes a
\texttt{get\_variables} method which can be used to retrieve the most recent
set of policy parameters. While in most cases the role of the variable source
is played by the \texttt{Learner} (introduced in Section~\ref{sec:learners}),
this has been generalized to include edge cases that may not yet be supported,
e.g.\ a sources that read parameters from disk, or some other form of caching.
Finally, rather than having each actor implement this polling procedure
directly we also provide a \texttt{VariableClient} object which handles
polling and other bookkeeping on a separate thread.

\begin{figure}
\centering
\includegraphics[width=0.8\textwidth, trim=6cm 5cm 5cm 4.4cm, clip]{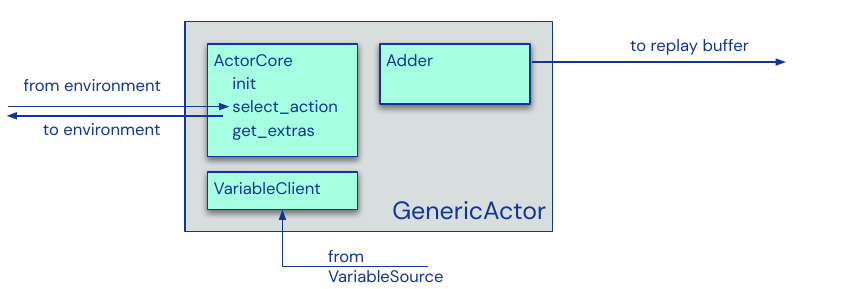}
\caption{An expanded view of the GenericActor and ActorCore implementation of
an Acme Actor.}
\label{fig:generic-actor-and-actor-core}
\end{figure}

While the \texttt{Actor} is a fundamental component of Acme, where possible we
prefer and recommend using the \texttt{GenericActor} object which implements
this interface; see Figure~\ref{fig:generic-actor-and-actor-core} for an
illustration. The \texttt{GenericActor} handles much of the boiler plate logic
that is most commonly necessary in distributed RL agents so that, as its name
suggests, it is generic enough to be used by most of our agents. This can be
done because the \texttt{GenericActor} defers the logic that differs between
agents to a component called the \texttt{ActorCore} which can generally be
thought of as the \emph{policy-specific} components of an actor. The
\texttt{ActorCore} is a simple container class, which holds the following
three pure functions:
\begin{itemize}
\item \texttt{init} produces a new initial state, usually at episode beginnings.
This state can hold, for example, the recurrent neural network (RNN) state of
agents endowed with memory (e.g.\ the R2D2 agent).

\item \texttt{select\_action} mirrors the corresponding \texttt{Actor} function
with the addition of state management.

\item \texttt{get\_extras} consumes the state and produces the \texttt{extras}
information that should be stored along with the corresponding timestep. This
is useful, for example, when learners require the log-probability of actions
taken in the environment for V-trace or Retrace corrections (e.g.\ Impala or
MPO).
\end{itemize}

\subsection{Experience replay and data storage}
\label{sec:replay}

Previous sections of this work have introduced actors which observe the data
generated from their interactions with the environment. An algorithm can
immediately consume that data in order to update its policy parameters or
store it for later consumption and possible re-evaluation. This distinction is
typically made when discussing on-policy algorithms in the first case and
off-policy algorithms in the second, where the storage of data for later use
is often referred to as \emph{experience replay} \citep{lin1992self,
schaul2015prioritized}. In this section we will discuss how Acme implements
both ends of the spectrum by making use of \emph{Reverb}
\citep{cassirer2021reverb}, a special-purpose storage system build for this
purpose.

Reverb is a high-throughput, in-memory storage system which is particularly
suited for the task of experience replay---especially in distributed settings.
In particular, Reverb exposes a client-server architecture where servers are
tasked with storing data and can be either co-located with the rest of the
agent infrastructure or run on a separate machine. Data can be written to the
server via a client interface at a per-timestep level and a subset of data
associated with one or more timesteps can be marked as sampleable units with
an associated priority. The ability to mark specific subsets of timestep data
as sampleable units enables one to efficiently expose different forms of data
required by specific algorithm implementations. Data is also organized into
different \emph{tables} which allows actors to write data to separate tables
which can then be individually sampled from. Reverb also defines multiple
mechanisms by which units can be sampled from replay including
first-in-first-out (FIFO), last-in-first-out (LIFO), and in proportion to
given priorities. Similar selection mechanisms are used to define which
elements should be removed from the replay when memory becomes full. For
prioritized sampling mechanisms, the priority associated with a sampleable
unit is a scalar value given when writing or updating the data to replay,
i.e.\ it is computed on both the actor and the learner respectively. By
marking units with a given priority (and later updating this priority) various
forms of prioritized replay \citep{schaul2015prioritized} can be implemented.

Reverb directly exposes a client-server architecture for writing and sampling
from replay, however in Acme it is typically not necessary to use this
interface directly. Data sampled from replay is exposed to Learners (described
in the next section) via a simple Python iterator where each iterate
corresponds to a batch of sampled data. Similarly, while Reverb's client
defines a low-level mechanism for writing data to replay, it is often simpler
to define that data semantically by making use of the \texttt{Adder}
interface. Adders implement an \texttt{add} method which consumes an action
$a_t$ and a timestep $\eta_t$, i.e.\ equivalent to the \texttt{observe} method
found in the actor interface. Instances of this object implement different
styles of pre-processing and aggregation of observational data that occurs
before insertion into the dataset. For example a given agent implementation
might rely on sampling transitions, $n$-step transitions, sequences
(overlapping or not), or entire episodes---all of which have existing adder
implementations in Acme. The different adder implementations take care of any
processing and data caching in order to write this data.

Although adders define \emph{what} data to write to replay, they do not
prescribe \emph{how} or at what \emph{rate} data is presented for consumption.
In a traditional synchronous learning loop, it is relatively easy to control
how many steps of acting in the environment an agent should perform between
each learning step. The ratio between acting and learning can, in turn, have a
dramatic effect not only on the sample efficiency of a given agent (i.e.\ the
number of environment steps required to reach a given performance) but also on
its long-term learning performance and stability. While the same is true for
distributed learning settings, the asynchronous nature of training agents in
this case makes it much more difficult to maintain a fixed ratio. Running the
learner and actor processes independently easily results in higher variance
which is often attributable to differences in the computational substrate
(e.g.\ different hardware and network connectivity) between processes but
pinpointing precise sources can be extremely challenging. Thus, maintaining a
fixed ratio of acting to learning steps in distributed setting comes at the
cost of either blocking the learner or actor processes and can be
computationally wasteful.

In Acme, these scaling issues can be mitigated through the use of Reverb's
\texttt{RateLimiter}. By adopting rate limitation, one can enforce a desired
relative rate of learning to acting, allowing the actor and learner processes
to run unblocked so long as they remain within some defined tolerance of the
prescribed rate. In an ideal setting, learner and actor processes are given
the correct proportion of computational resource to run unblocked by the rate
limiter. However if due to network issues, insufficient resources, or
otherwise, one of the processes starts lagging behind the other, the rate
limiter will block the latter while the former catches up. While this may
waste computational resource by keeping the latter idle, it does so only for
as long as is necessary to ensure the relative rate of learning to acting
stays within tolerance. Rate limitation can also be used to simulate a queue
data structure which allows one to implement more purely online algorithms in
conjunction with FIFO samplers/removers; in Section~\ref{sec:agents} we will
highlight a number of algorithms that make use of this system.

\subsection{Learners}
\label{sec:learners}

Learners in Acme are the components which implement parameter updates given
data. Often this data is gathered by actors and sampled from replay as
introduced in the previous sections, however as we will see in
Section~\ref{sec:offline} this interface can be completely reused for static
datasets containing offline data. Learners also have the most general
interface, due largely to the fact that they must accommodate the greatest
amount of variation between different algorithm implementations. This
interface consists largely of a \texttt{step} function which performs a single
learning step updating the agent's parameters---typically, although not
exclusively, corresponding to a step of gradient descent. Learners also expose
a \texttt{get\_variables} method which returns the current state of its
parameters. By defining this method Learners implement the
\texttt{VariableSource} interface introduced when describing actors, and this
method is primarily used to update actors as they interact with the
environment.

In practice learner implementations should also be initialized using a dataset
iterator. Although this iterator can be backed by experience replay, that is
not a requirement and in fact it is precisely this design choice that greatly
simplifies the process of defining offline algorithms. The offline setting for
reinforcement learning, as introduced in Section~\ref{sec:modern-rl}, can be
seen as an extreme version of off-policy learning in that such algorithms rely
on a fixed dataset and no additional data generation. What this means is that
any algorithm implemented via the \texttt{Learner} interface can be applied in
the offline setting purely by eliminating the actor/replay part of their
algorithm. Similarly many algorithms in the LfD setting can be constructed by
adapting online algorithms and specially mixing online and offline data.
However, while this means that any Acme algorithm can be easily adapted to
offline settings this certainly does not mean that they will perform well---in
fact most algorithms built for such settings have to be carefully designed to
avoid over-generalization as we will highlight in Section~\ref{sec:agents}.

Finally, while different algorithms implemented in Acme require different data
formats (e.g.\ trajectories, sequences, etc.) for offline baselines and
experiments, we generally rely on standard mechanisms for providing this data.
In the online setting this is provided by the relevant Adder implementations,
and in the offline setting the necessary formats are defined by the RL
Datasets (RLDS) package~\citep{ramos2021rlds}. This package stores data as
full episodes, and provides utilities to transform them into the desired
trajectory format.

\subsection{Defining agents using Builders}
\label{sec:builders}

As described at the beginning of this section, an RL agent is defined by its
combination of actors, learners, and replay (although as we will see in
Section~\ref{sec:offline} the actor may only be needed for evaluation and
replay may be replaced by a static dataset). While these components can be
used individually, in Acme we also provide general purpose mechanisms for
running \emph{experiments} which instantiate the agent and its environment and
run its learning process. We will detail this in the following sections,
however in order to do so we need a full definition of the agent. In Acme this
is accomplished by agent \emph{builder} objects which define how to construct
the individual components.

The builder abstraction encapsulates the full definition of an agent algorithm
in a way which allows applying those algorithms to a wide variety of learning
tasks represented as environments. This means that each algorithm should make
as few assumptions as possible about the environment they are interacting with
(e.g. some environments provide image observations, others provide an encoding
of the internal environment state). To achieve this, all Acme algorithms can
be configured with an \emph{environment spec} and a \emph{network module}. The
environment spec object provides a lightweight specification of the types and
shapes of environment inputs and outputs---knowledge required by many of the
agent's components. The network module object defines the neural network
architecture required to construct and learn the agent's policy. Specifying
networks outside of the builder allows for reusing a single agent
implementation for a wide variety of different environments. To simplify the
process of training an agent from a user perspective, most algorithms provide
a means to construct a default set of networks given an environment spec.

With that out of the way we can define the interface for builder objects, which 
includes the following functions:
\begin{itemize}
\item \texttt{make\_policy} constructs the policy as a function of the
environment spec and networks.

\item \texttt{make\_replay\_tables} returns a list of replay tables, i.e.\ the
Reverb collections to which data will be written; this allows adders to write
data differently between multiple tables which can then be sampled on the
learner side. As the format of the tables may depend on the shape of the inputs 
and outputs of the environment, as well as the outputs of the policy, this 
function accepts both the environment spec and policy as inputs.

\item \texttt{make\_adder} constructs the adder which will be used to record
data to replay. This is a function of both the environment spec and networks
for the same reasons as given above. It is also a function of the replay
client which will be constructed from the replay tables; this allows low-level
access necessary for the adder to write its data.

\item \texttt{make\_actor} returns an actor as a function of the policy, an
adder, and a variable source; these are the necessary components to evaluate
actions, add data, and synchronize parameters. Also passed are the environment
spec which is necessary for initialization and a random key for seeding any
stochasticity in the policy.

\item \texttt{make\_learner} returns the learner object which handles all
parameter updates for the agent. This is primarily a function of the dataset
iterator and network modules, however the environment spec is also necessary
for initialization, and a random key is also provided to seed any randomness
in the learning process.

\item \texttt{make\_dataset\_iterator} constructs the dataset iterator as
a function of the replay client.

\end{itemize}
Additionally, to allow for logging and diagnostic capabilities the learner is
also equipped with \emph{shared counter} and \emph{logger} objects which
respectively allow for maintaining metadata counts shared among the different
agent components and logging this and other metadata; we will describe loggers
in more detail in the following section.

The builder definition contains everything necessary to define an agent and
situate it within an environment loop. While this can be done manually it is
generally more useful to run a predefined \emph{experiment} as we will
describe in the next section. Ultimately, as we will show in
Section~\ref{sec:distributed} we can use the same interface to instantiate and
run a distributed experiment. This also allows for some level of algorithmic
composition (as we will see in Section~\ref{sec:imitation}) where higher-level
algorithms are defined by composing builders. In the following section we will
show how this interface can be used to construct and run an agent; see also
Listing~\ref{lst:run-experiment}.

\subsection{Running an agent}
\label{sec:running}

In Acme we refer to the combination of a builder, network definitions, and
environment as an \emph{experiment}. While the concept of an experiment is
somewhat loosely defined, the components necessary to run an experiment can be
concretely specified using the \texttt{ExperimentConfig} datatype. This
datatype provides factories for network and environment instances along with a
builder instance; note also that the builder itself can be seen as a
collection of factories for lower-level modules. Given such a config, Acme
provides a simple \texttt{run\_experiment} function which will instantiate all
of the necessary actor, learner, environment loop components, etc.
Listing~\ref{lst:run-experiment} shows a simplified version of this function
which summarizes how all of the previously introduced components are connected
together.

\begin{figure}
\includegraphics[width=\textwidth, trim=2cm 5cm 2cm 5cm, clip]{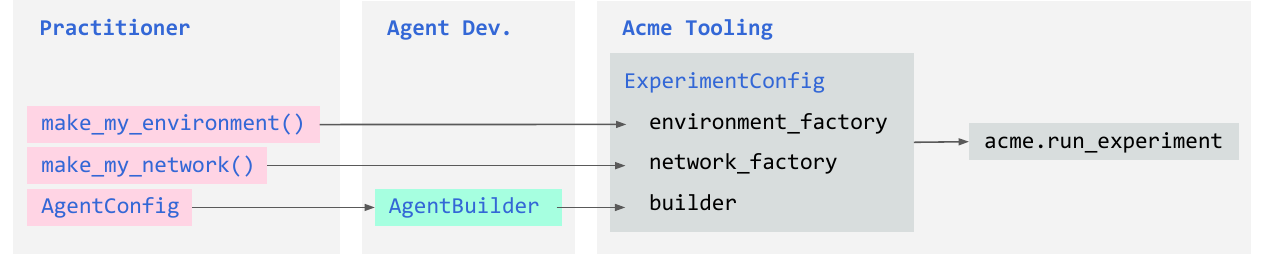}
\caption{This diagram illustrates how experiments are configured and
run in the Acme framework.}
\label{fig:config-and-run}
\end{figure}

Figure~\ref{fig:config-and-run} breaks down the different elements necessary
to configure and run an experiment. In particular we have also loosely split
these components into different levels at which they sit within Acme's
framework. At the lowest (left-most) level is the tooling provided by Acme
which defines the configuration and the \texttt{run\_experiment} function;
from a practitioner or agent developer perspective these should be usable
as-is. Going up one level, a user who is interested in developing a new
algorithm can do so by creating a new agent builder, which may require
modifying or creating new learners, actors, etc.\ (which are in turn
constructed by the builder). However, from the perspective of a practitioner
who is interested in using an algorithm for a novel environment or network
architecture they need only specify environment and network factories and any
agent-specific configuration.

\begin{algorithm}[t!]
\begin{algorithmic}[1]
\caption{Pseudocode for a simplified version of \texttt{run\_experiment}. 
This requires environment and network factories as well as a builder specifying
the agent. The pseudocode below will then construct and run the agent in
an environment loop. For simplicity this omits logging operations as well as
some simple helper functions. Also not shown is a \texttt{\_make\_learning\_actor}
wrapper which is necessary for running learning operations, but for a full
overview one should refer to the code. \vspace{0.3em}}
\label{lst:run-experiment}
\REQUIRE{\texttt{environment\_factory}, \\
         \hspace{2.81em}\texttt{network\_factory}, \\
         \hspace{2.81em}\texttt{builder}.}
\item[]{\vspace{0.8em}\textit{\# Create an environment and get a spec describing inputs/outputs.}}
\STATE{\texttt{environment $\gets$ environment\_factory()}}
\STATE{\texttt{environment\_spec $\gets$ \_make\_environment\_spec(environment)}}

\item[]{\vspace{0.8em}\textit{\# Create the networks and use this to build a policy
                                 which generates actions.}}
\STATE{\texttt{networks $\gets$ network\_factory()}}
\STATE{\texttt{policy $\gets$ builder.make\_policy(environment\_spec, networks)}}

\item[]{\vspace{0.8em}\textit{\# Create replay and the dataset.}}
\STATE{\texttt{replay\_tables $\gets$ builder.make\_replay\_tables(environment\_spec, policy)}}
\STATE{\texttt{replay $\gets$ \_make\_replay(replay\_tables)}} 
\STATE{\texttt{dataset $\gets$ builder.make\_dataset\_iterator(replay)}}

\item[]{\vspace{0.8em}\textit{\# Create the learner which updates parameters.}}
\STATE{\texttt{learner $\gets$ builder.make\_learner(environment\_spec, networks, dataset, replay)}}

\item[]{\vspace{0.8em}\textit{\# Create an adder (to add data) and an actor (to generate data).}}
\STATE{\texttt{adder $\gets$ builder.make\_adder(replay, environment\_spec, policy)}}
\STATE{\texttt{actor $\gets$ builder.make\_actor(environment\_spec, policy, learner, adder)}}

\item[]{\vspace{0.8em}\textit{\# Wrap the actor to periodically run training steps on the learner.}}
\STATE{\texttt{actor $\gets$ \_make\_learning\_actor(actor, learner, ...)}}

\item[]{\vspace{0.8em}\textit{\# Create the main loop and run it.}}
\STATE{\texttt{environment\_loop $\gets$ EnvironmentLoop(environment, actor)}}
\STATE{\texttt{environment\_loop.run()}}
\end{algorithmic}
\end{algorithm}

Experiments can additionally contain a \emph{logger} factory which controls
where and how performance and debugging data is written. Loggers, which also
allow for rate control and aggregation, are simple objects that implement a
\texttt{write} method taking a dictionary of numerical data. These objects are
then passed to the learner and to any created environment loops.
While loggers control where and how to record data they say nothing about what
data should be recorded. In the case of any loggers passed to a learner
object, the learner directly controls what data is recorded. However, when
using a generic actor and environment loop it may be desirable to record
additional data that is environment-specific. To account for this, each
experiment can additionally be given an \emph{observer} factory which will be
constructed and passed to any created environment loops. This shares a similar
interface to the actor and allows a user to record additional data that can
then be written by the logger. Importantly this is purely a user-side callback
and requires no additional change to the algorithm (i.e.\ the builder).

Another feature typically required of an experiment in Acme is the ability to
periodically evaluate the agent's performance during training. This problem of
evaluation is related but distinct from both the losses optimized by the
learner and the performance of the experience-collecting actors. The losses
are often just a proxy for the actual performance whereas the actors are
typically exploratory in some manner and will take under-explored actions to
achieve this at the cost of degraded performance. Acme provides a default
evaluation mechanism that instantiates an environment loop and actor/policy
that runs in parallel with training. Such actors do not record their
experience so as to not affect the learning process. The policy factory
associated with the agent is allowed to take a Boolean-valued parameter
\texttt{eval} which allows configuring them for the needs of the evaluation
(e.g. turn off exploration in the policy). In addition, experiments can be
given generic \emph{evaluator} factories which allow users to define and
compute custom evaluation metrics specific to their experiments (e.g.\ to
evaluate the distribution of a given policy).

\subsection{Distributed agents}
\label{sec:distributed}

\begin{figure}
\centering
\includegraphics[width=0.7\textwidth, trim=2cm 4cm 10cm 2cm, clip]{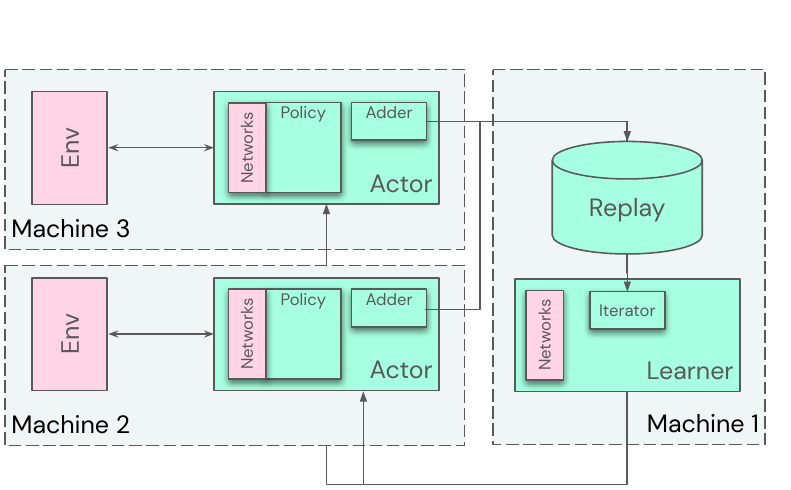}
\caption{This diagram illustrates a distributed training loop. Here two
environment loops are running on different machines, feeding data to a single
replay, which is co-located with the Learner process on a third machine. }
\label{fig:environment-loop-distributed}
\end{figure}

A major, ongoing aspect of RL research is the question of how to scale these
algorithms up in terms of the amount of data that can be both produced and
consumed. A particularly successful---and general---strategy for addressing
this question relies on generating data asynchronously by making use of
distributed data generation, typically by interacting with multiple
environments in parallel~\citep[see e.g.][]{nair2015massively, mnih2016a3c,
horgan2018apex, barthmaron2018d4pg}. This has two benefits: the first being
that environment interactions can occur asynchronously with the learning
process, i.e.\ we allow the learning process to proceed as quickly as possible
regardless of the speed of data gathering. The second benefit is that by
making use of more actors in parallel we can accelerate the data generation
process. See Figure~\ref{fig:environment-loop-distributed} for an illustration
of a distributed agent with two data generation processes.

In previous sections we have already introduced the primary split in Acme
agents between acting, learning, and storage components. While this was
primarily introduced through the lens of composability, it serves another
crucial purpose: namely, by running these components in different threads,
processes, or in particular on different machines we can implement a
distributed agent using all of the same agent building blocks. To this end
Acme exposes a \texttt{make\_distributed\_experiment} function which roughly
shares the same interface as \texttt{run\_experiment} introduced in the
previous section, albeit with some additional control over
distributed-specific parameters such as the number of data-generating actors.
Rather than running the experiment, this function returns a \emph{program}
data structure that describes the distributed computation and how to launch
it. This datatype is defined by \emph{Launchpad} \citep{yang2021launchpad},
the package we use to build and launch these programs. In what follows we will
briefly describe Launchpad and its use within Acme.

Launchpad provides a programming model that simplifies the process of defining
and launching instances of distributed computation. The fundamental concept of
this model is that of a Launchpad \emph{program} which represents computation
as a directed graph of service \emph{nodes}. Edges in this graph denote
communication between nodes which are made via remote procedure calls. By
making the graph representation explicit Launchpad makes it easy to both
design and later modify a program's topology in a single, centralized way.
Launchpad also provides a \texttt{launch} function which takes a program, a
\emph{launch type}, and potentially a collection of \emph{resource types}. The
launch type can be used to specify the distribution mechanism---e.g.
multi-processing, multi-threading, distributed machines, etc. Similarly the
resource specification can be used to specify resource requirements such CPU,
RAM, etc.\ where required by the launch type. This model is also extensible to
different launch platforms, in particular one platform specifically targeted
by Launchpad is that of Google Cloud.

In Acme our use of Launchpad is entirely contained within the
\texttt{make\_distributed\_experiment} function introduced earlier. This
method is tasked with constructing the program and adding all of the
associated nodes. The Launchpad nodes in our case correspond exactly to the
top-level components introduced in the previous section(s)---i.e.\ the
environment loops (containing the actor and environment), learners (which
perform the update loop), and the Reverb service itself. The program datatype
is then returned by this function and launching the distributed experiment can
be performed by using Launchpad's \texttt{launch} mechanism directly. Note
that in the previous section we briefly alluded to a \emph{learning actor}
which controlled the rate at which learner steps are taken as a function of
the data gathered by the actor(s). In the distributed case this is no longer
necessary as access to the dataset is purely asynchronous and use of a
rate-limiter via Reverb handles this rate naturally.

Given this mechanism, combined with the particular split between components
that we have chosen in Acme, results in a system that largely divorces the
question of how to run (and scale) an algorithm from what to run. Many
performance optimizations can be implemented directly in the background during
program construction (i.e.\ the \texttt{make\_distributed\_experiment}
function) allowing for algorithms to be developed without worrying about these
optimizations. In the following section we introduce a number of agents built
with Acme and describe at a high level how these are constructed.

\subsection{Modifications for offline agents}
\label{sec:offline}

Finally, alternate experiment \emph{runners} can be exposed for defining this
loop in different problem settings. For example, Acme also includes a
\texttt{run\_offline\_experiment} function which runs a simplified experiment
with no data-generation processes (i.e.\ the actor and adder components) and a
fixed dataset. Note that this can---and often does---make use of an
environment for evaluation purposes \citep{paine2020hyperparameter}, although
as noted above this can be replaced with an arbitrary evaluation mechanism. As
noted earlier in Section~\ref{sec:learners}, this process can make use of an
OfflineBuilder which doesn't define components such as the adder. However,
this function also takes as input agents defined using the general
\texttt{Builder} interface, allowing for all algorithms to be used in the
offline setting. In a nutshell, implementing an offline agent amounts to
implementing a learner and using the offline versions of the builder, the
experiment config and the runner. Therefore, any online agent can readily be
used in an offline setup.

\section{Agent Algorithms}
\label{sec:agents}

In this section, we describe a collection of agents which have been
implemented and included with Acme. Our intention is for these agents to serve
both as clear and succinct \emph{reference implementations} of their
respective RL algorithms as well as strong research \emph{baselines} in their
own right. We are periodically running regression tests to ensure their
performance stays stable.

As described in Section~\ref{sec:acme} the implementation of an agent
comprises the entire apparatus to both collect new data as well as continually
learn improved policies from that data. For many of the agents we introduce,
however, their primary differences will be contained within the
\emph{learner}, i.e.\ the module which consumes data and updates the agent's
policy parameters.

In Section~\ref{sec:agent-preliminaries} we will first give a brief
presentation of a number of common principles shared across agents implemented
in Acme. Then, we will describe the individual agents, focusing primarily on
the major distinctive elements of each particular agents. The agents have also
been separated based on their primary problem setting---as introduced in
Section~\ref{sec:modern-rl}---and we have labeled each agent with tags (e.g.\
\tagoffpolicy) to highlight their specific characteristics. While the tags we
use should be well-explained by the given agent descriptions, they are
detailed more clearly in Appendix~\ref{sec:tag-lookup}. For historical reasons
we will also highlight agents for which distributed data generation was a
primary contribution of the original work. However, beyond the distinction
between on- and off-policy data generation this is not a significant
algorithmic distinction in Acme, as all algorithms built using Acme's modular
interfaces can readily be run in a distributed fashion. Finally, note that
while this collection of agents is meant to serve both as a snapshot of
currently implemented agents as well as a broad overview of Acme's
capabilities, it is not meant to be comprehensive.

\subsection{Preliminaries}
\label{sec:agent-preliminaries}

As introduced by Equation~\eqref{eq:return} in Section~\ref{sec:modern-rl},
the objective of nearly all of the agents we will cover is to maximize the
\emph{expected return} $\E_\pi\big[G_t\big] = \E_\pi\big[\sum_{i \geq 0}
\gamma^{i} R_{t+i}\big]$ with respect to their policy $\pi$. As noted in the
introduction the expectation is explicitly dependent on the policy $\pi$,
however we have left implicit all the other distributions involved which arise
from the environment. This is important because $G_t$ is a function of entire
trajectories which depends on both the stochasticity of the policy and the
environment. However, in what follows we will leave these additional
distributions implicit and will introduce more detail as necessary.

One common construction used to optimize the expected return is a \emph{value
function} which writes the return as a function of the current state of the
system. This can either be phrased as a \emph{state-action value function}
$Q^\pi(s_t,a_t)=\E_\pi[G_t|s_t,a_t]$, also known as the \emph{Q-function}, or
as a a state-only version $V^\pi(s_t)=\E_\pi[G_t|s_t]$. Also frequently used
is the \emph{advantage} function $Q_\pi^\text{adv}(s_t,
a_t)=Q^\pi(s_t,a_t)-V^\pi(s_t)$ which denotes the advantage gained by taking
action $a_t$ rather than following the policy $\pi$. Note that in this section
we have returned to the fully-observable MDP setting i.e.\ where the states
$s_t=o_t$ are equivalent to the observations. This is purely to simplify
notation due to the fact that the Markov assumption is necessary to derive
many of the following equations. In practice, and as noted in
Section~\ref{sec:modern-rl}, such methods can be applied directly to the
partially-observable setting (while sacrificing any performance guarantees) or
by utilizing recurrent networks to approximate the fully history of
observations. Later in this section we will highlight specific algorithms that
make use of recurrence to attack partial observability.

While estimates of the value function can be used directly without explicit
parameterization---often leading to so-called \emph{Monte Carlo}
methods---they can also be defined recursively, i.e.\ with the Q-function
being written as
\begin{align}
    Q^{\pi}(s_t, a_t)
    = &\ \E_\pi[R_t + \gamma Q^\pi(S_{t+1}, A_{t+1}) \mid s_t, a_t \big].
    \label{eq:bellman-equation}
\end{align}
This construction is commonly referred to as the \emph{Bellman
equation}~\citep{bellman1957mdp} and a similar approach can also be used for
$V^\pi$. Note again that the expectation is with respect to any stochasticity
in the reward and dynamics models where the subsequent action $A_{t+1}$ is
drawn according to the distribution $\pi(\cdot|S_{t+1})$. If the policy is
deterministic the action selection can be written more simply as
$A_{t+1}=\pi(S_{t+1})$. By examining the recursive formulation in
Eq.~\eqref{eq:bellman-equation} we can see that the true value function
associated with any given policy $\pi$ is a fixed-point of the Bellman
equation. Solving this recursion naively with function approximators, however,
can prove to be unstable~\citep{baird1995residual}. To avoid this instability,
a common approach is to compute the right-hand-side of the equation using a
\emph{previous} estimate of the value function---held fixed for a certain
window of time---and use this as a target for the left-hand-side. This can be
formulated as minimization of the loss function
\begin{align}
    L(\phi) &=
    \tfrac12 \E_{\rho_\pi}
    \big[
    (Q_\phi(S_t, A_t) - Y_t )^2
    \big]
    \text{ where } 
    Y_t = R_t + \gamma \E_{A\sim\pi(\cdot|S_{t+1})}\big[Q_{\phi'}(S_{t+1}, A)\big]
    \label{eq:td-error}
\end{align}
Here outer expectation is taken with respect to the visitation distribution
$S_t,A_t,R_t,S_{t+1}\sim\rho_\pi(\cdot)$ of transitions under the policy $\pi$
and the expectation of the target $Y_t$ is taken with respect to the policy
$\pi$. This loss is also known as the Temporal Difference (TD)
error~\citep{sutton2018reinforcement}. In the context of deep RL the network
associated with parameters $\phi$, i.e.\ those that are being optimized, are
referred to as the \emph{online network} while the \emph{target network} is
associated with the fixed set of parameters $\phi'$. This entire process is
also known as \emph{bootstrapping} due to the way in which an older version of
the value function is used to to recursively improve an estimate of itself.
The use of bootstrapping also allows us to eliminate expectations of entire
trajectories and instead focus on individual transitions.

Above we have distinguished two ends of the spectrum defined by direct and
recursive value function formulations, numerous methods exist between these
two ends \citep[see e.g.][]{schulman2016high} and are fully supported in Acme.
One simple technique that is used frequently relies on the $n$-step TD error;
this generalizes the TD error and simply replaces the single reward in the
bootstrap target $Y_t$ with a discounted sum of $n$ rewards and the following
state and action pair with $(S_{t+n}, A_{t+n})$. This behavior is supported in
Acme by the choice of Reverb adder, either by storing length $n$ sequences or
using the more efficient $n$-step transition adder.

Given a value estimate, we can then turn to the question of how to use this to
improve the agent's policy. One special, albeit very common case is that of
finite, discrete action spaces in which the Q-function can be used to
implicitly parameterize the policy as $\pi(s_t)=\argmax_a Q(s_t, a)$. Since
every change to the Q-function also updates the policy, this can be
incorporated directly into the target from Eq.~\eqref{eq:td-error} as
\begin{align}
    Y_t &= R_t + Q_{\phi'}(S_{t+1}, \argmax_a Q_{\phi'}(S_{t+1}, a)).
\end{align}
In other settings the policy can instead be explicitly parameterized, written
as $\pi_\theta$, and these parameters can be adjusted in order to maximize the
expected return $J(\theta)=\E_{\pi_\theta}[G_0]$. Classically, such algorithms
rely on a stochastic policy with actions of the form
$a_t\sim\pi_\theta(\cdot|s_t)$; in such cases the gradient of $J$, also known
as the policy gradient~\citep{sutton2000policy}, can be written as
\begin{equation}
    \nabla_\theta J(\theta)
    = \E_{\rho_{\theta'},\pi_{\theta'}}[
    Q^{\pi_{\theta'}}(S, A)\nabla_\theta\log\pi_\theta(A|S)].
    \label{eq:policy-gradient}
\end{equation}
Here $\pi_{\theta'}$ represents the policy under a previous set of parameters
and $\rho_{\theta'}$ represents the stationary distribution of states under
that policy. Here we have explicitly separated the distributions over states
and actions as in practice the approximate distribution of states may rely on
a staler set of states (e.g.\ the case of replay). Additionally, while we have
written the policy gradient using the true value function $Q^{\pi_{\theta'}}$
in practice this quantity is typically not available and will need to be
approximated. The simplest such approximation relies on replacing the
Q-function with a sum of rewards sampled from the underlying system (the
distribution $\rho_{\theta'}$ can also be rewritten as a distribution over
full episodes for which individual state-action pairs can be summed over).
Such \emph{vanilla} policy gradient algorithms are also typically
\emph{on-policy} due to the fact that the data used to estimate the Q-function
is generated from the same policy that is being optimized. Alternatively, we
can use an explicit Q-function estimate as described earlier. Such algorithms,
typically referred to as \emph{actor-critic}, estimate both the policy (actor)
and the value function (critic) where each iteration typically corresponds to
a single step of both components. Note that the use of the term actor should
not be confused with the more general, but related concept of an actor which
we have introduced previously and which it predates. Finally, it is also
possible to compute the policy gradient for deterministic policies
$a_t=\pi_\theta(s_t)$ where the policy gradient in this case takes the form
\begin{equation}
    \nabla_\theta J(\theta)
    = \E_{\rho_{\theta'}}[\nabla_a Q_{\phi'}(S, a)\nabla_\theta\pi_\theta(S)].
\end{equation}
This estimate was introduced by \citep{silver2014dpg} who also showed that it
is the limiting case of the stochastic policy gradient as the policy variance
tends to zero.

Updates for the value function and policy gradient computations rely closely
on the way in which data is presented to them. On-policy algorithms---typified
by direct value estimates---require that the policy used to generate data be
the same, or very close, to that which is being optimized. As a result,
on-policy algorithms typically make use of a queue for data storage so that
data can be processed as soon as it is produced (and often discarded
thereafter). However, the recursive formulation often results in off-policy
algorithms for which data generation need not follow the same policy as that
which is being optimized. Experience replay allows for random access to the
generated data and also exposes an additional knob to tune: the \emph{behavior
policy}, i.e.\ the policy used to generate data and fill replay. These
resulting algorithms rely on sampling from replay often as an approximation to
the steady state distribution of the policy being optimized. However, since
this distribution cannot be represented exactly it is also common to make use
of \emph{prioritized experience replay} wherein the probability of each sample
is reweighted by some \emph{priority}, typically in proportion to their TD
error \citep{schaul2015prioritized}.

Additionally, while many of the standard tools and machinery of supervised
learning can be directly applied in an RL context, frequently these techniques
must be adapted to the fact that in RL the supervisory signal (long-term
value) is only indirectly observed via the immediate rewards. One such
technique that has seen a recent resurgence in interest is the use of
distributional value functions, i.e.\ value functions which estimate the
entire distribution of expected return rather than only its expectation
\citep{bellemare2017distributional,dabney2018iqn,barthmaron2018d4pg}; this
results in a modified version of the TD error introduced earlier. The network
architecture used for representing the value function can also play an
important role in the performance of an agent. This is shown most clearly in
Dueling DQN~\citep{wang2015dueling} which learns a separate state-only value
function and an \emph{advantage} function (representing the
advantage/disadvantage of a single immediate action over the expected value)
all with a shared embedding. One technique that is particularly unique to RL
is typified by the use of separate networks for action selection and computing
the target value function when estimating the Q-function. This technique,
known as Double Q-Learning is frequently used to avoid overestimation bias;
see \cite{hasselt2010double} for more details. Finally, the use of
\emph{regularization} is also just as commonly used RL as it is in supervised
learning. In an RL context this often takes the form of an additional penalty
defined by the Kullback-Leibler (KL) divergence between the agent's policy and
some reference distribution. For example, previous iterates of the policy
(often the target parameters) can be used as a reference. This constrains the
magnitude of allowable updates
\citep{abdolmaleki2018mpo,levine2018inference,vieillard2020munchausen} in
order to stabilize the learning process.

\subsection{Online RL}

\paragraph{DQN} \tagdiscrete \tagoffpolicy \tagQnetwork \tagbootstrapping 
\\[\parskip]
Deep Q-Networks (DQN)~\citep{mnih2013playing, mnih2015atari} is a modern
variation on the classical Q-Learning algorithm~\citep{watkins1992}, which
uses a deep neural network to estimate the Q-function in discrete action
spaces. DQN learns $Q_\phi$ by iteratively applying the recursive update
described in the previous section and calculated by minimizing the loss in
Equation~\eqref{eq:td-error} (with actions selected by the greedy policy). One
optimization implemented by this algorithm relies on the fact that that it is
specifically targeted at discrete action spaces. Rather than implement
$Q_\phi$ as a function of both observations and actions, DQN actually learns a
vector-valued function of the observations where the $a$-th output is the
value of the associated action, i.e.\ $Q_\phi(s,a)=[Q_\phi(s)]_a$ . This
entire vector can then be calculated using a single pass through the network.
In terms of data-generation DQN relies on an $\epsilon$-greedy behavior policy
for exploration, i.e.\ one in which with probability $\epsilon<1$ a uniformly
random action is selected and otherwise the greedy, maximum valued action is
taken. As done in many other algorithms the value of $\epsilon$ is frequently
annealed towards zero as learning proceeds. Data generated by this process is
added to replay as $n$-step transitions which are used to form $n$-step
returns during learning. In our implementation of DQN, and following in the
spirit of Rainbow DQN \citep{hessel2018rainbow}, we also include a number of
recent enhancements including Double Q-Learning, dueling networks, and
prioritized experience replay using the TD error as the priority.

\paragraph{Munchausen DQN} \tagdiscrete \tagoffpolicy \tagQnetwork
\tagbootstrapping \\[\parskip]
Munchausen DQN (MDQN)~\citep{vieillard2020munchausen} is an explicitly
entropy-regularized and implicitly KL-regularized version of DQN. In order to
apply this regularization MDQN uses a stochastic policy given by a softmax of
the Q-function, $\pi_\phi(a_t| s_t)\propto \exp (\tfrac1\tau Q_\phi(s_t,
a_t))$ with temperature $\tau$; note that we have written the policy
$\pi_\phi$ due to the fact that it is indirectly parameterized by $\phi$ via
the Q-function. The target from Eq.~\eqref{eq:td-error} can then be modified
to add the log-policy (scaled by $\alpha \in (0,1)$) to the reward function
and replace the argmax policy by expectation over the softmax, resulting in
\begin{align*}
    Y_t = R_t 
    + \gamma 
    \E_{A\sim\pi_{\phi'}(\cdot|S_{t+1})}
    \big[Q_{\phi'}(S_{t+1}, A) - 
    {\color{BrickRed} \tau \log\pi_{\phi'}(A| S_{t+1})}\big]
    + {\color{BrickRed} \alpha \tau \log\pi_{\phi'}(A_t| S_t)}.
\end{align*}
Note that the additional regularization terms have been highlighted in
{\color{BrickRed}red} and that when $\tau \rightarrow0$ we recover Q-learning. This implicitly regularizes the policy towards the
previous one and widens the action-gap. Note that in the above equation the
target $Y_t$ is a random variable whose expectation we will take with respect
to the state, action, and next-state.

\paragraph{R2D2} \tagdiscrete \tagoffpolicy \tagQnetwork \tagbootstrapping
\\[\parskip] 
Recurrent Replay Distributed DQN (R2D2) is a distributed, recurrent extension
of DQN. This involves a Q-function computed by a recurrent network, which is
arguably better adapted to partially observable environments. From an
implementation perspective this replaces the feed-forward networks used in DQN
with recurrent networks, adding full sequences (rather than transitions) to
replay, and adapting the learning process to deal with these trajectories. For
arbitrary trajectories sampled from replay, the initial state of the recurrent
network can be set by replaying some initial set of states that will not be
used for learning, referred to as \textit{burn-in}. R2D2 additionally makes
use of an invertible value rescaling to enhance robustness to different reward
scales; for more details see~\citet{pohlen2018observe}.

\paragraph{D4PG}  \tagcontinuous \tagoffpolicy \tagQnetwork \tagPnetwork
\tagbootstrapping \\[\parskip]
Distributed Distributional Deterministic Policy Gradients
(D4PG)~\citep{barthmaron2018d4pg} is an off-policy, actor-critic algorithm
adapted for continuous action spaces. In particular D4PG is a distributed
extension of the Deep Deterministic Policy Gradient (DDPG) approach
of~\citet{lillicrap2016ddpg}, which also adds a distributional critic and
$n$-step transitions. The original implementation of this work makes use of a
Q-function represented by a categorical distribution
\citep{bellemare2017distributional}, although a mixture-of-Gaussians critic
was also experimented with (see also MPO, introduced later). D4PG makes use of
$n$-step returns, again implemented via the $n$-step transition adder. The
other major contribution of the original work was its use of distributed
experience generation, however as all agents in Acme are distributed this is
not a major axis of variation with other algorithms. Note that in our later
experiments we will also utilize a single-actor, non-distributed version of
D4PG to simplify comparison with the other continuous control methods.

\paragraph{TD3}  \tagcontinuous \tagoffpolicy \tagQnetwork \tagPnetwork 
\tagbootstrapping \\[\parskip]
The Twin Delayed DDPG (TD3) algorithm~\citep{fujimoto2018addressing} also
extends DDPG with three primary additions. All of these modifications are
restricted to the algorithm's computation of loss and as a result strictly
occur on the learner. First, this algorithm utilizes a technique similar to
Double Q-learning, which uses two estimates of $Q_\phi$ in its target to avoid
overestimation, however the variation introduced by this work always uses the
smaller of the two estimates. Second, TD3 introduces a delay under which the
policy $\pi_\theta$ is updated less frequently than the Q-function during
optimization. This is an attempt to delay policy updates until the value
estimate has converged. Finally, TD3 also introduces a modification to the
critic estimate where noise is added to actions calculated at subsequent
states in order to smooth the target estimate in an effort to avoid policies
which exploit errors in the Q-function.

\paragraph{SAC} \tagcontinuous \tagoffpolicy \tagQnetwork \tagPnetwork \tagbootstrapping \\[\parskip]
Soft Actor-Critic (SAC)~\citep{haarnoja2018soft, haarnoja2018softapplications}
is an off-policy, actor-critic algorithm that optimizes an entropy regularized
objective. Much like MDQN this approach requires a stochastic policy, however
unlike MDQN (which this algorithm predates) SAC utilizes a separate
parameterized policy $\pi_\theta$ that outputs the parameters of some
underlying distribution, often a Gaussian much like D4PG and TD3. The critic
objective is derived from Equation~\eqref{eq:td-error} where the target
includes an additional entropy term, i.e.
\begin{align}
    Y_t
    &= R_{t} + \gamma \E_{A \sim \pi_{\theta'}\left(\cdot | S_{t+1} \right)}\left[Q_{\phi'}(S_{t+1}, A) 
    - \alpha \log \pi_{\theta'}(A |S_{t+1})\right]
\end{align}
and where the objective used to optimize the policy similarly corresponds to
the expected return regularized by the log-policy,
\begin{align}
    J(\theta) 
    = \E_{S\sim\rho_{\theta'}, A\sim\pi_{\theta}\left(\cdot | S \right)} 
    \big[Q_{\phi'}(S, A) - \alpha \log \pi_\theta (A |S) \big].
\end{align}
The amount of regularization is controlled by a temperature $\alpha$ which can
be either set or learned. Finally, it is worth highlighting that SAC makes use
of the \emph{reparameterization trick} to parameterize its stochastic policy
as a deterministic policy and independent noise.

\paragraph{MPO} \tagcontinuous \tagoffpolicy \tagQnetwork \tagPnetwork \tagbootstrapping \\[\parskip]
Introduced by~\citet{abdolmaleki2018mpo}, Maximum a posteriori Policy
Optimization (MPO) is another actor-critic algorithm which employs
KL-regularization in the policy optimization step. Due to the fact that MPO is
agnostic to the choice of how its Q-function is learned our implementation
does so in an equivalent manner to DDPG. Its main contribution, instead, lies
in terms of how its policy is learned. In the literature it is often
formulated as under the guise of \textit{RL as
inference}~\citep{levine2018inference} and is solved using the classical
expectation-maximization (EM) algorithm. Because of this particular approach,
the objective minimized by gradient descent takes the following peculiar form:
\begin{align}
    J_{\eta, \alpha}(\theta)
    &= 
    \E_{S \sim \rho_{\theta'}} \left[
    \E_{A \sim \pi_{\theta'}\left(\cdot |S\right)} \left[
            \exp \left( \tfrac1\eta Q_{\phi'}(S,A) \right)
            \log\pi_{\theta}(A|S)
        \right]
        - \alpha
        D_{\mathrm{KL}}(\pi_{\theta'}(\cdot|S)\ \|\ \pi_{\theta}(\cdot|S))
        \right]
\end{align}
where the second term is the KL divergence regularization. Optimization of
this objective corresponds to the M-step of the EM algorithm, and the
exponentiated Q-function results from solving the E-step analytically. This
regularization ensures that the online policy $\pi_\theta$ does not move too
far from the target network $\pi_{\theta'}$ all while encouraging the online
policy to continue adapting if necessary. Finally, the $\eta$ and $\alpha$
terms are not hyperparameters, and are instead \emph{dual variables} with
losses of their own allowing them to be adaptively adjusted to the changing
data distribution. In Acme, and the included experiments below, we also
include Distributional MPO (DMPO) which makes use of a distributional critic
as in D4PG and mixture-of-Gaussians MPO (MOG-MPO) which utilizes a MOG
distribution for this critic.

\paragraph{PPO} \tagdiscretecontinuous \tagonpolicy \tagPnetwork \tagVnetwork
\tagmc \tagbootstrapping \\[\parskip]
Proximal Policy Optimization (PPO) \citep{schulman2017proximal} is an
on-policy agent that maximizes a simplified version of the objective proposed
by the Trust Region Policy Optimization (TRPO) algorithm
\citep{schulman2015trpo}. Whereas TRPO explicitly constrains its updates by
limiting the KL divergence between consecutive policies, PPO instead enforces
these constraints using a proximal operator. These updates can be expressed as
\begin{align}
    J(\theta) &=
    \E_{S \sim \rho_{\theta}, A \sim \pi_{\theta}\left(\cdot | S\right)} 
    \Big[ \min\Big(
    \frac{\pi_\theta(A|S)}{\pi_{\theta'}(A|S)} Q^\text{adv}_\phi(S,A), 
    \text{clip}\Big(
    \frac{\pi_\theta(A|S)}{\pi_{\theta'}(A|S)}, 1-\epsilon, 1+\epsilon\Big)
    Q^\text{adv}_\phi(S,A)\Big) 
    \Big] \label{eq:ppo_objective}.
\end{align}
Here the $Q^\text{adv}_\phi(S,A)$ term is a Generalized Advantage Estimate
\citep{schulman2016high} which is related to the standard Q-function and
expresses the advantage that action $A$ has over the expected value of the
given policy. An estimate of this advantage is derived by mixing direct
Monte-Carlo return estimates with a state-value function estimate (learned
through Bellman updates) similarly to TD($\lambda$). The term $\epsilon$ is a
hyperparameter that controls how much the new policy $\pi_\theta$ can deviate
from the past iterate $\pi_{\theta'}$.

\paragraph{IMPALA} \tagdiscrete \tagonpolicy \tagPnetwork \tagmc
\tagbootstrapping \\[\parskip] 
IMPALA~\citep{espeholt2018impala} is an actor-critic algorithm that uses an
importance correction mechanism called V-trace to learn from on-policy and
potentially off-policy data. The $n$-step V-trace TD target for values is
\begin{align}
    Y_t = V_{\phi'}(S_t) + \sum_i^{n-1} \gamma^i \Big(\prod_k^{i - 1} c_{t + k}\Big)
    \delta_{t + i} (R_{t+i} + \gamma V_{\phi'}(S_{t + i + 1}) - V_{\phi'}(S_{t
    + i})),
\end{align}
with $\delta_t = \min(\bar{\delta}, \frac{\pi(A_t | S_t)}{\mu(A_t | S_t)})$
and $c_t = \min(\bar{c}, \frac{\pi(A_t | S_t)}{\mu(A_t | S_t)})$ truncated
importance sampling weights. It uses a distributed set of actors that write
sequences of interaction data in a queue with FIFO insertion order that the
learner gets updates from.

\subsection{Offline RL}

\paragraph{BC} \tagdiscretecontinuous  \tagoffline \tagPnetwork  \\[\parskip]
Behavior Cloning (BC) \citep{pomerleau1991efficient} is a supervised learning
method for offline RL and imitation learning which does not require online
interactions with the environment. Given an offline dataset of interactions
with the system and a parameterized policy BC minimizes the error between the
policy's actions, conditional on the recorded state, and the recorded action.
Additionally, this approach is quite general-purpose and acts more like a
family of algorithms based on the distance metric used. In particular BC can
make use of a regression loss (e.g.\ for continuous control) or a
classification loss (e.g.\ for  discrete control).

\paragraph{TD3 + BC} \tagcontinuous \tagoffline \tagQnetwork \tagPnetwork
\tagbootstrapping \\[\parskip]
Owing to their off-policy nature, algorithms such as TD3 can directly support
offline RL due to the fact that they learn from interactions not generated by
the current version of the policy. However, the lack of feedback from the
environment usually leads to divergence at least partly because of the
extrapolation of actions not taken in states from the offline dataset
$\mathcal{D}$. Adding a BC regularization term \citep{fujimoto2021minimalist}
in the actor objective helps mitigate this phenomenon
\begin{align}
  J(\theta) &= 
  \E_{S, A \sim \mathcal{D}}
  \Big[ Q_\phi(S, \pi_\theta(S)) - \lambda \|A - \pi_\theta(S)\|^2 \Big].
\end{align}

\paragraph{CQL} \tagdiscretecontinuous \tagoffline \tagQnetwork \tagPnetwork \tagbootstrapping \\[\parskip]
Similarly to TD3 + BC, Conservative Q-learning (CQL) attempts to prevent
action extrapolation by using a regularization term when estimating the
Q-function \citep{kumar2020conservative}. In particular it does so by adding
an additional term to the critic loss which amounts to a penalty on deviations
of the critic's estimate under an alternative reference policy $\mu$. This
results in a loss that combines the standard TD-error with this additional
penalty
\begin{align}
    L(\phi) = 
    \E_{S, A, R, S' \sim \mathcal{D}}
    \Big[ 
    \E_{\hat A \sim \pi_\theta(\cdot | S')} \big[(R + \gamma Q_{\phi'}(S', \hat A) - Q_\phi(S, A))^2 
    \big] +
    \alpha\big(\E_{\hat A \sim \mu\left(\cdot | S\right)}\big[Q_\phi(S, \hat A)\big] - Q_\phi(S, A)\big)
    \Big].
\end{align}
In practice the reference measure $\mu$ can be chosen in a min-max fashion
where typical values result in it being proportional to the exponential of the
Q-function (and thus the penalty term becomes a log-sum-exp of the
Q-function). For more details see the referenced work.

\paragraph{CRR} \tagcontinuous \tagoffline \tagQnetwork \tagPnetwork 
\tagbootstrapping \\[\parskip]
Critic-Regularized Regression (CRR)~\citep{wang2020critic} is a modification of behavior cloning that
restricts training to those example transitions whose advantage (i.e.\ the
difference between the Q- and V-functions) is positive. Here the Q-function
can be trained by simply minimizing the TD-error (e.g.\ as in TD3) and the
V-function can be written as the expected value of the Q-function evaluated
under the current policy. As a result the policy updates can be written to
maximize the log-probability of an observed transition only if this advantage
is positive, i.e.\
\begin{align}\label{eq:crr}
    J(\theta)
    = \E_{S,A \sim \mathcal{D}} 
    \Big[ \log\pi_\theta(A\vert S)
    \,\mathbb 1_{>0}\big[Q(S, A) - \E_{\pi_{\theta'}}[Q(S, A')] \big]
    \Big]
\end{align}
where $\mathbb 1_{>0}$ is an indicator function which takes on a value of one
when its argument is greater than zero (and is zero otherwise). Note that
different variants of CRR exist that correspond to replacing the indicator
function with and exponential $\exp(\cdot/\beta)$ whose sharpness is
controlled by the temperature $\beta$. Additionally the inner expectation can
be replaced with a maximum over the Q-function with samples taken from the
policy.

\paragraph{One-Step CRR} \tagcontinuous \tagoffline \tagQnetwork \tagPnetwork 
\tagbootstrapping \\[\parskip]
While the above methods attempt to learn a Q-function that can be used for
policy improvement, they must contend with potential extrapolation and
over-estimation of actions not present in the dataset. An alternative strategy
is to completely avoid evaluating the critic on any other policy than the
observed one as suggested by \citet{brandfonbrener2021offline}. The resulting
critic is used to separately improve a learned policy, via similar techniques
to that employed by CRR. This corresponds to simply replacing actions sampled
from the policy with actions sampled directly from the replay buffer, i.e. by
modifying the inner expectation from CRR.

\paragraph{BVE} \tagdiscrete \tagoffline \tagQnetwork \tagbootstrapping \\[\parskip]
Behavior value estimation (BVE) \citep{gulcehre2021regularized} is a one-step
policy improvement algorithm where the agent, instead of trying to recover the
optimal policy, learns a Q-function that estimates the behavior value
$Q^{\beta}(s, a)$  from the dataset. To achieve this, BVE eliminates the
max-operator from the target network when computing the Bellman updates for
the TD-learning. The behavior values can be learned with the SARSA tuples or
directly predict the Monte Carlo returns in the offline RL with discrete
actions. For example
\begin{align*}
    L(\phi) &= 
    \E_{S, A, R, S' \sim \mathcal{D}}
    \big[ 
   (R + \gamma Q_{\phi'}(S', A^\prime) - Q^\beta_\phi(S, A))^2 \big] \text{ or}, \\
    &= \E_{S, A, R, S' \sim \mathcal{D}}
    \big[ 
   (\sum_i \gamma^i R_i - Q^\beta_\phi(S, A))^2 \big].
\end{align*}
\citet{gulcehre2021regularized} found that for Atari, during the training,
just using the SARSA tuples with will bootstrapping and doing a single step of
policy improvement when the agent is evaluated in the environment works the
best, and this is the setup we use in this paper.

\paragraph{REM} \tagdiscrete \tagoffline \tagQnetwork \tagbootstrapping \\[\parskip]
Random ensemble mixtures (REM) \citep{agarwal2019optimistic} is an offline-RL
agent that makes use of ensemble of Q-functions parameterized by neural
networks to estimate the Q-values.  The key insight behind REM is that the
convex combination of multiple Q-value provides a robust estimate of the
Q-value
\begin{align*}
    L(\phi) = 
    \E_{S, A, R, S' \sim \mathcal{D}}
    \Big[ 
    \E_{\alpha \sim P_\Delta} \big[(R + \gamma \max_{a} 
    \sum_{k} \alpha_k Q^k_{\phi'}(S^\prime, a) - \sum_{k} \alpha_k Q^k_\phi(S, A))^2 
    \big]
    \Big].
\end{align*}
where $P_\Delta$ represents the probability distribution over a $(K-1)$
simplex with $\Delta^{K-1} = \{ \alpha \in \mathbb{R}^K: \sum_{i=1}^K a_i = 1,
\alpha_k \ge 0, \forall k \in [1, K]\}$. During the training $\alpha$'s are
sampled from uniform distribution such that $\alpha_k \sim U(0, 1)$ and then
normalized $\alpha_k = \frac{\alpha_k}{\sum_i \alpha_i}$ to be in the $(K-1)$
simplex . For action-selection in the environment, REM uses the average of the
Q-value estimates, i.e $Q(s,a) = \sum_k^K Q_{\phi}^k (s,a) / K$.

\subsection{Imitation Learning}
\label{sec:imitation}

In this section, we describe Imitation Learning algorithms which include a
direct RL method. The tags are left blank as the overall algorithm inherits
properties from its direct RL method.

\paragraph{AIL} \tagdiscretecontinuous   \\[\parskip]
Adversarial Imitation Learning (AIL) is a framework introduced by
\citet{ho2016generative} that follows the two-player game paradigm set forth
by Generative Adversarial Networks (GANs). In this setting an agent (the
generator) is trained to produce trajectories that are indistinguishable from
demonstrations as determined by a trained discriminator. The objective
consists of matching the state-action occupancy measure of the policy, denoted
$d_{\pi_\theta}$, with that of an expert, denoted $d_e$. This matching is done
by minimizing their Jensen-Shannon divergence between the two occupancy
measures \citep{ho2016generative,ke2019imitation}. In practice, the policy is
trained with an online RL algorithm whose reward is given by a classifier
which tells whether a transition comes from the demonstration data or the
agent interaction data. Letting $D(s, a)$ denote the probability of
classifying a given state-action pair as expert-generated, the policy can be
trained via the base RL algorithm with a reward of $r(s, a) = -\log
(1-D_\omega(s,a))$. The discriminator itself can be trained by maximizing the
following objective
\begin{equation*}
K(\omega)
=
\E_{O, A \sim d_{\pi_\theta}} [\log D_\omega(O, A)] - 
\E_{O, A \sim d_e} [\log(1-D_\theta(O,A))]
.
\end{equation*}
A number of variants of GAIL have been proposed and have shown to improve
performance using an off-policy agent rather than an on-policy agent
\citep{kostrikov2018discriminator}, using regularization in the discriminator
\citep{miyato2018spectral,gulrajani2017improved} or using different
formulation of the rewards \citep{fu2018learning}. We refer the reader to
\citet{orsini2021matters} for an extensive review of the different variants
and their influence on performance. In the rest of the paper, the GAIL variant
studied uses TD3 as the direct RL method similarly to DAC
\citep{kostrikov2018discriminator} with the original GAIL reward.

\paragraph{SQIL} \tagdiscretecontinuous  \\[\parskip]
Soft Q Imitation Learning \citep{reddy2019sqil} is a method based on a limit
case of AIL, since it consists in using a constant positive reward if the
transition comes from the expert demonstrations and a zero reward if the
transition comes from the agent interactions. The agent learns from balanced
batches between expert demonstrations and agent interactions using an
off-policy online RL agent (in our study, the agent used is SAC).

\paragraph{PWIL} \tagcontinuous \\[\parskip]
Primal Wasserstein Imitation Learning \citep{dadashi2020primal} is a
non-adversarial imitation learning method that aims at minimizing the primal
formulation of the Wasserstein distance between the state-action occupancy
measure of the agent and the one of the expert. The reward is derived from an
upper bound of the Wasserstein distance. In the following, the direct RL agent
considered is D4PG.

\subsection{Learning from Demonstrations} 

\paragraph{SACfD/TD3fD} \tagcontinuous \tagoffpolicy \tagQnetwork
\tagPnetwork \tagbootstrapping  \\[\parskip]
We propose two natural modifications to the SAC agent and the TD3 agent
compatible with the Learning from Demonstrations setting and inspired from
\citet{Vecerik:etal:arXiv17}. In this setting, contrary to the Imitation
Learning setting, the environment has a well-defined reward function and the
demonstrations come with environment rewards. We can thus modify any
off-policy agent (in this case, SAC and TD3) to sample batches of transitions
either from environment interactions or from demonstration data with a ratio
that is a hyperparameter.

\section{Experiments}
\label{sec:experiments}

In order to showcase Acme agents, we provide in this section a set of
experiments from multiple agents tackling the settings described in
Section~\ref{sec:modern-rl}. This further highlights the flexibility of our
framework due to the fact that a number of agents can be evaluated in multiple
setups, e.g.\ TD3 alone can be used online, offline by making use of TD3+BC,
AIL can be used with TD3 as its direct RL agent, and TD3fD can be used to
learn from demonstrations. Note however that while the learning curves are
representative and should be informative they are not meant to establish a
strict hierarchy between the different agents since these could be further
improved by tweaking hyperparameters on a per-environment basis. Instead these
results should demonstrate that the agents are implemented correctly and are
able to reproduce the published baselines. Additionally, one can reproduce
these baselines by simply using the runfiles in the \texttt{baselines/} folder
of Acme. Similarly, all hyperparameters used the experiment can be read
directly in these provided scripts.

\subsection{Protocol} \label{protocol}
For each setup, we evaluate agents on a common set of tasks that are standard
in the literature (e.g. Atari games for discrete control) for a fixed number
of interactions with the environment. Most agents can have two policies: a
\textit{training} policy that interacts with the environment while balancing
exploration and exploitation, and an \textit{evaluation} policy that is fully
exploitative.  Each experimental setup will specify which policy is used in
the corresponding figures. For each evaluated agent, we performed a
hyperparameter search for a configuration that is reasonably consistent with
performance reported in the literature across all tasks. Agent performance is
computed across multiple random seeds (specified for each setup), and episode
returns are grouped in 50 contiguous equal size windows of steps (actor steps
for online algorithms and learner steps for offline algorithms). The figures
below show the resulting average and standard deviation. The standard
deviation captures both the inter-seed and intra-seed variability. In the
presented results, a single hyperparameter configuration is run across the
full setup. One can thus tweak further the hyperparameter of each algorithm
for a specific task and reach much higher performance.

\subsection{Environments}

\paragraph{Arcade Learning Environment}
The Arcade Learning Environment (ALE) \citep{bellemare2013ale} provides a
simulator for Atari 2600 games. ALE is one of the most commonly used benchmark
environments for RL research. The action space ranges between 4 to 18 discrete
actions (joystick controls) depending on the game. The observation space
consists of 210 x 160 RGB image. We use a representative subset of five Atari
games to broadcast the performance of our discrete agents. We also use several
pre-processing methods on the Atari frames including giving zero-discount on
life loss, action repeats with frame pooling, greyscaling, rescaling to 84x84,
reward clipping, and observation stacking, following
\citet{machado2018revisiting}.
\paragraph{DeepMind Control suite} The DeepMind Control suite
\citep{tassa2018controlsuite} provides a set of continuous control tasks in
MuJoCo \citep{todorov2012mujoco} and has been widely used as a benchmark to
assess performance of continuous control algorithms. The tasks vary from the
simple control problems with a single degree of freedom (DOF) such as the
cartpole and pendulum, to the control of complex multi-joint bodies such as
the humanoid (21 DOF). The DeepMind Control suite enables to control an agent
both in a feature-based observation space and a raw pixel-based observation
space. The following results are limited to the former.
\paragraph{Gym environments} The Gym suite \citep{brockman2016gym} provides a
wide range of continuous control tasks in MuJoCo and is a testbed for
continuous control algorithms. In the following, we focus on locomotion tasks,
that share similarities with the ones defined in the DeepMind Control suite.
We also consider the Adroit manipulation tasks \citep{rajeswaran2017learning},
a collection of 4 tasks with a 24-DOF simulated hand.
\begin{figure}[hbt!]
    \centering
    \includegraphics[width=0.9\textwidth]{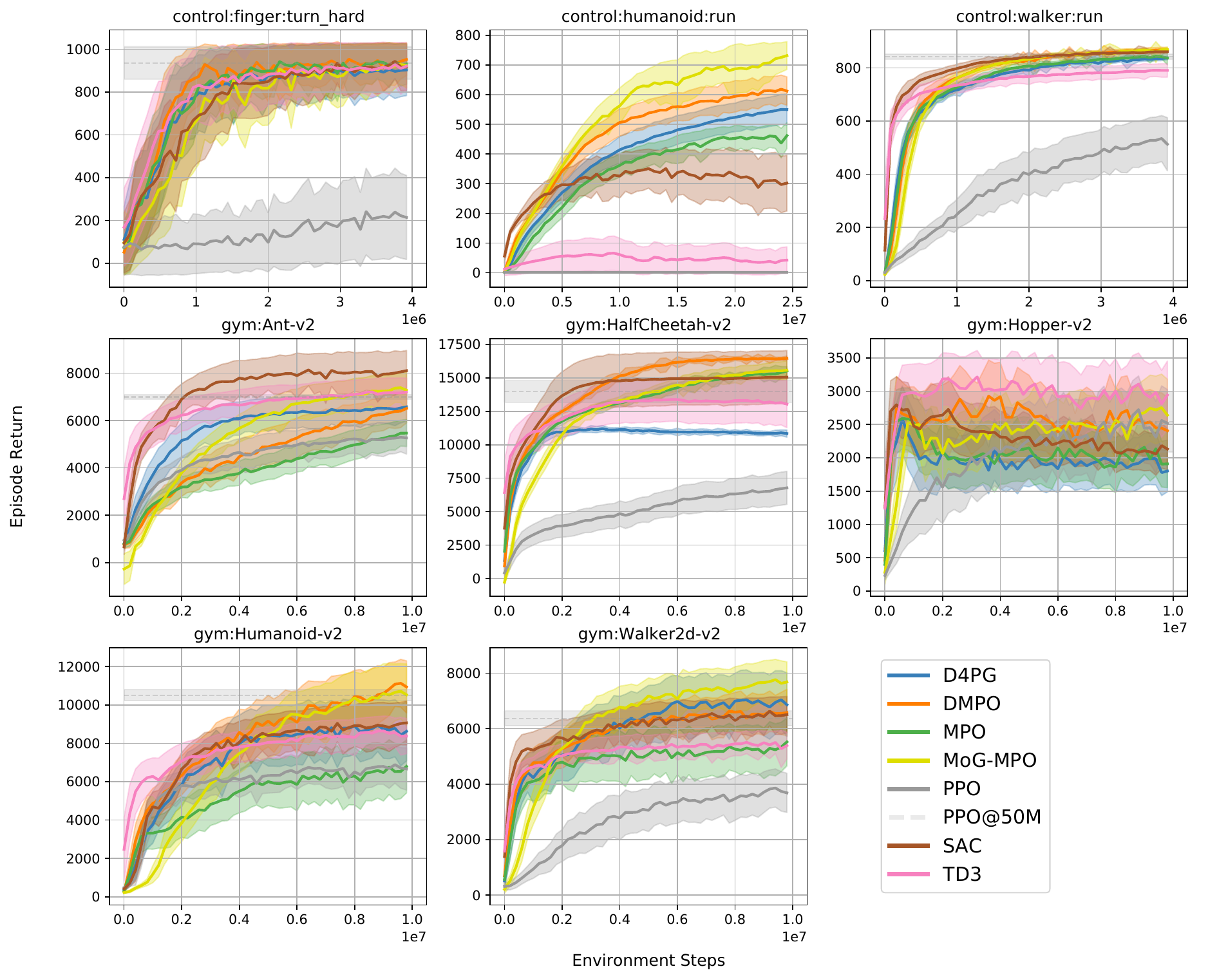}
    \caption{\textbf{Continuous Control.}  Episode returns obtained by the
    evaluation policies. Here we plot the means and standard deviations over
    25 seeds. Gym environments are run for 10 million environment steps, DM
    Control environments for 4 million, except for DM Control humanoid which
    is run for 25 million steps.}
    \label{fig:xp:crl}
\end{figure}
\pagebreak
\subsection{Results}
\paragraph{Continuous Control}
We evaluate PPO, TD3, SAC, D4PG and MPO (as well as MPO variants) on the
environments described in Section \ref{protocol}. Policy and value networks
have a common three hidden layer architecture with 256 units.  Hyperparameters
are kept constant across environments for each algorithm, except for D4PG for
which we change the support of the distributional critic for Gym and DM
Control. We report the results in Figure~\ref{fig:xp:crl}, where we can
observe the average episode return for the evaluation policy (without
exploration noise) as a function of the environment steps. For the sake of
completeness, we also include the results of PPO at convergence as a dashed
horizontal line. To obtain this value, we run the exact same configuration of
PPO with 256 actors for 50M environment steps and plot the average and
standard deviation of the episode return obtained between 45M and 50M steps,
where the algorithm has converged.

\begin{figure}[htb!]
    \centering
    \includegraphics[width=0.9\textwidth]{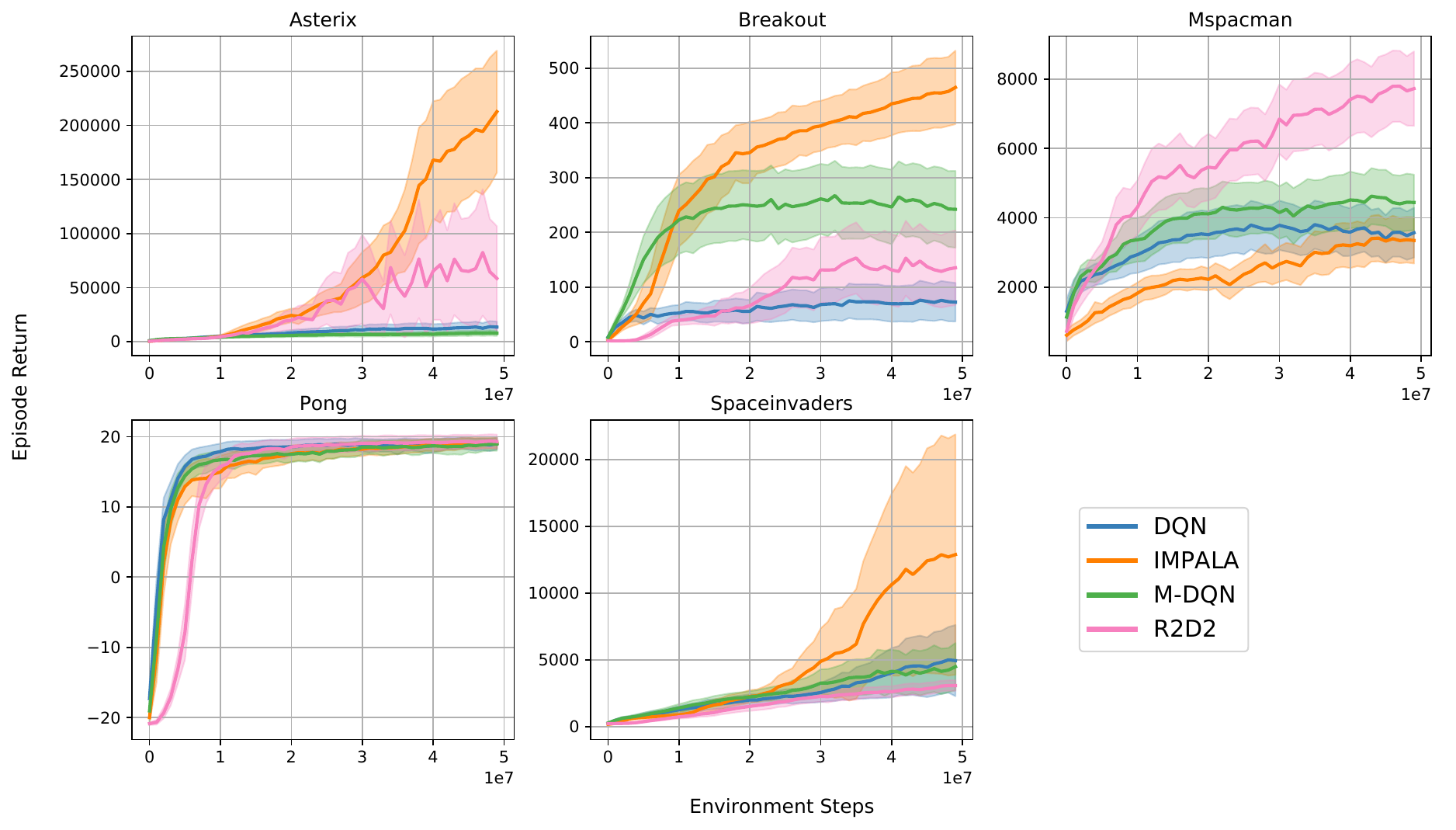}
    \caption{\textbf{Discrete Control.} Episode returns obtained by the
    training policies. Here we plot the means and standard deviations over 5
    seeds. Each environment is run for 50 million environment steps, which
    corresponds to 200 million frames, due to the agent's actions being
    repeated four times on the environment side~\citep{mnih2013playing}.}
    \label{fig:xp:drl}
\end{figure}

\paragraph{Discrete Control}
We evaluate DQN, R2D2, Impala and Munchausen-DQN on 5 Atari games:
\texttt{Asterix}, \texttt{Breakout}, \texttt{MsPacman}, \texttt{Pong} and
\texttt{SpaceInvaders}. For DQN and M-DQN, both algorithms use the same
network, the orginal three convolutional layer followed by one dense layer
from \citet{mnih2013playing}. For R2D2 and Impala, the network architecture is
recurrent. Note that both DQN and M-DQN use Adam as their optimizer, which
considerably improves results compared to RMSProp. The results depicted in
Figure~\ref{fig:xp:drl} are computed across 5 random seeds, in a distributed
setup.
Although the unified setup (distributed actors, sticky actions, zero-discount
on life loss, 200M frames) makes it harder to compare to all specific
reference papers, performance of the resulting policies overall match the
results provided in the literature.
In the chosen 200M frames regime, which uses an order of magnitude less
samples than in the original papers of recurrent agents (R2D2/IMPALA),
recurrent architectures still seem to help with performance.

\begin{figure}[htb!]
    \centering
    \includegraphics[width=0.9\textwidth]{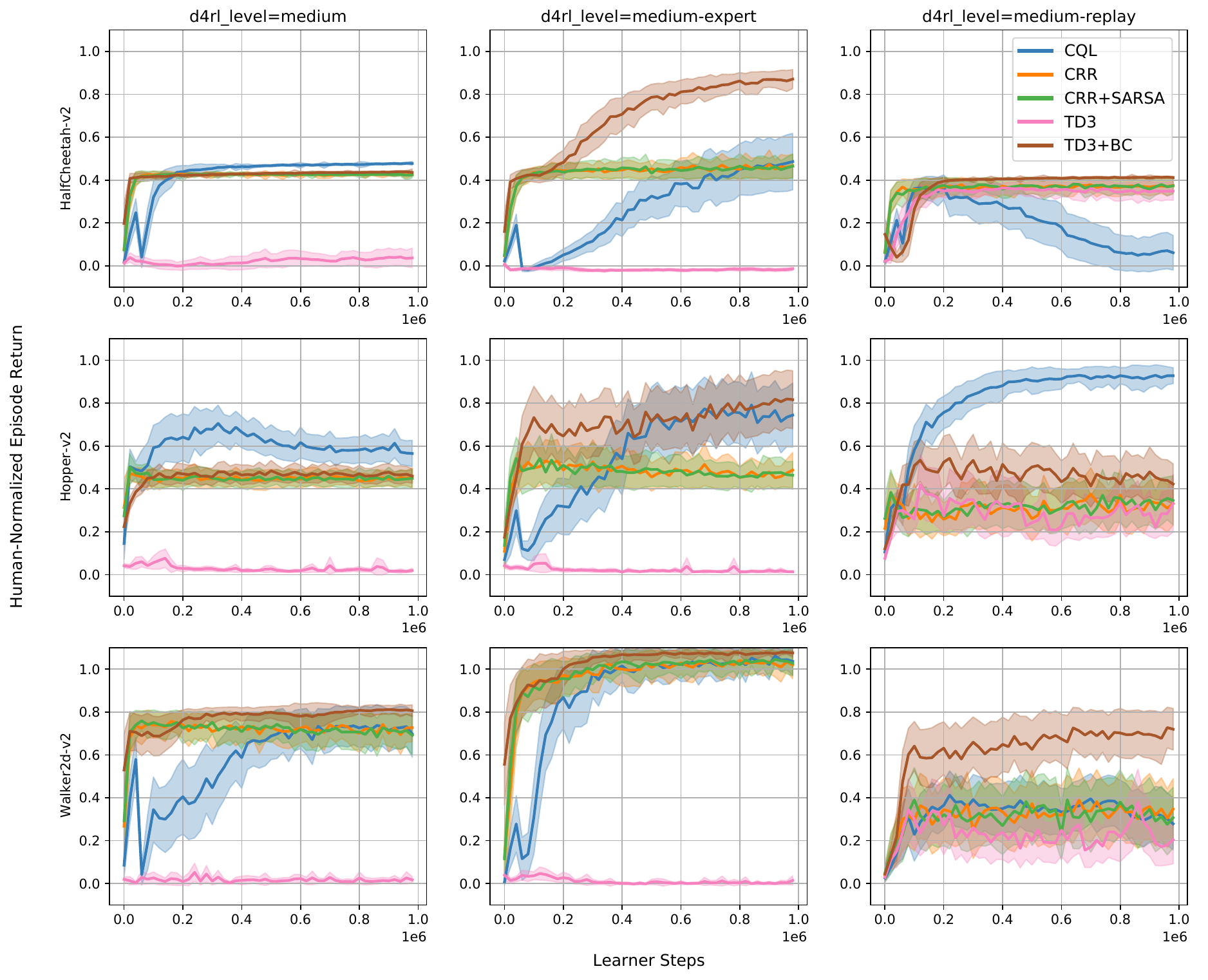}
    \caption{\textbf{Offline RL.} Mean and standard deviation of the episode
    return obtained by the evaluation policies over 25 seeds in 1M learner
    steps. Each column is a different D4RL level. Each row is a different
    environment.}
    \label{fig:xp:orl}
\end{figure}

\begin{figure}[htb!]
    \centering
    \includegraphics[width=0.9\textwidth]{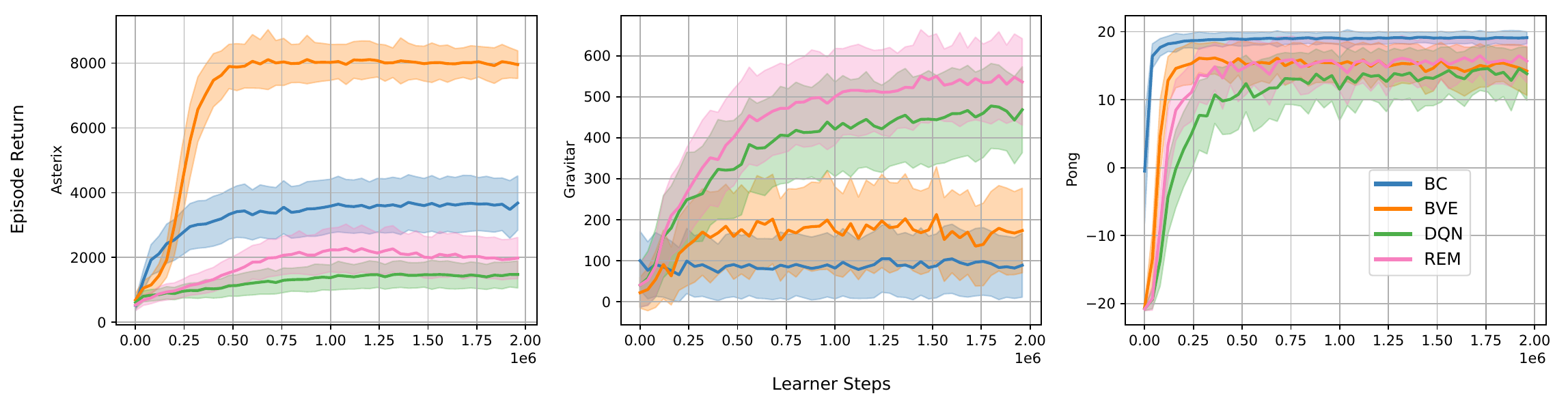}
    \caption{\textbf{Offline RL for Atari.} Mean and standard deviation of the
    episode return obtained by the evaluating policies over 5 seeds in 2M
    learner steps for BVE, REM, DQN and BC on 3 Atari games.}
    \label{fig:xp:orl_atari}
\end{figure}

\pagebreak
\paragraph{Offline Reinforcement Learning} We evaluate TD3, TD3+BC, CQL, CRR
and CRR+SARSA on the Gym MuJoCo \citep{brockman2016gym} locomotion
environments mentioned in Section~\ref{protocol} with D4RL datasets
\citep{fu2020d4rl}. The D4RL datasets is a collection of datasets generated by
agents trained online. Figure~\ref{fig:xp:orl} compares the performance of
these algorithms on different D4RL levels (medium-expert, medium,
medium-replay). For CQL, we observed that an initial pre-training of the
policy for 50K steps significantly improves the performance.
We also trained and evaluated BC, BVE, REM, and offline DQN on offline Atari
datasets from RL Unplugged offline RL benchmark suite \citep{gulcehre2020rl}
with the best hyperparameters reported in the paper. We noticed in Figure
\ref{fig:xp:orl_atari} that on dense reward games, such as Pong and Asterix,
the one-step RL agent BVE performs better than REM and DQN. However, on
Gravitar, a sparse reward game, the methods that do multiple steps of policy
improvement, such as DQN and REM, perform better than BVE.

\begin{figure}[htb!]
    \centering
    \includegraphics[width=0.9\textwidth]{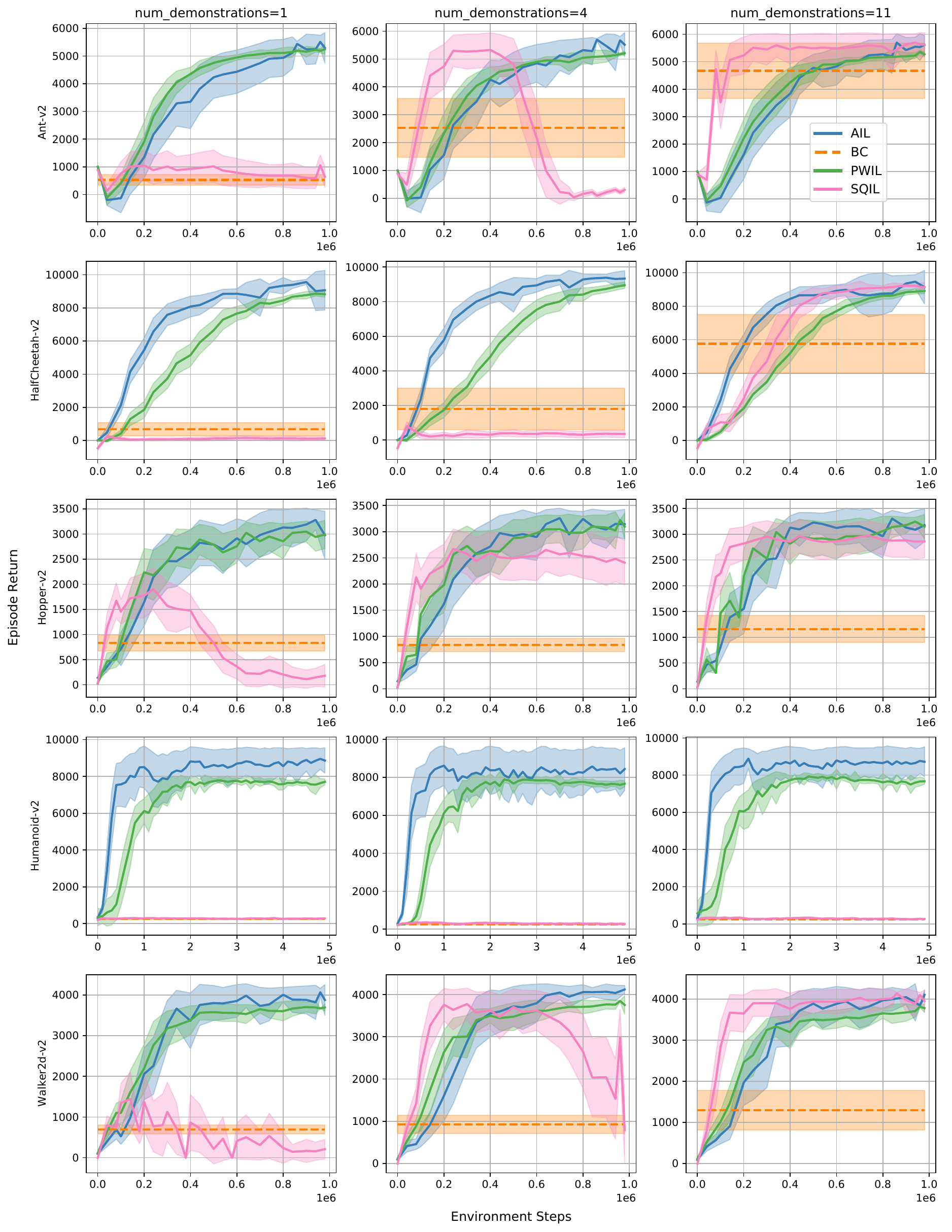}
    \caption{\textbf{Imitation Learning.} Episode return obtained by the evaluation policies over 25 seeds
    in 1M environment steps. Each column corresponds to a different number of
    demonstration trajectories. Each row is a different environment. The BC
    algorithm, which is trained offline, is depicted as an dashed line.}
    \label{fig:xp:il}
\end{figure}

\pagebreak
\paragraph{Imitation Learning} We evaluate AIL, PWIL, SQIL and BC on the
locomotion environments mentioned in Section~\ref{protocol}. We use the
demonstrations generated by \citet{orsini2021matters} which consists in 50
trajectories of a SAC agent trained to convergence on the original reward of
the environment. For each seed, a fixed number of demonstration trajectories
is sampled ($1$, $4$ or $11$); we did not subsample the trajectories. Figure
\ref{fig:xp:il} indicates that the performance of all methods improves with
the number of demonstrations.

\begin{figure}[htb!]
    \centering
    \includegraphics[width=0.8\textwidth]{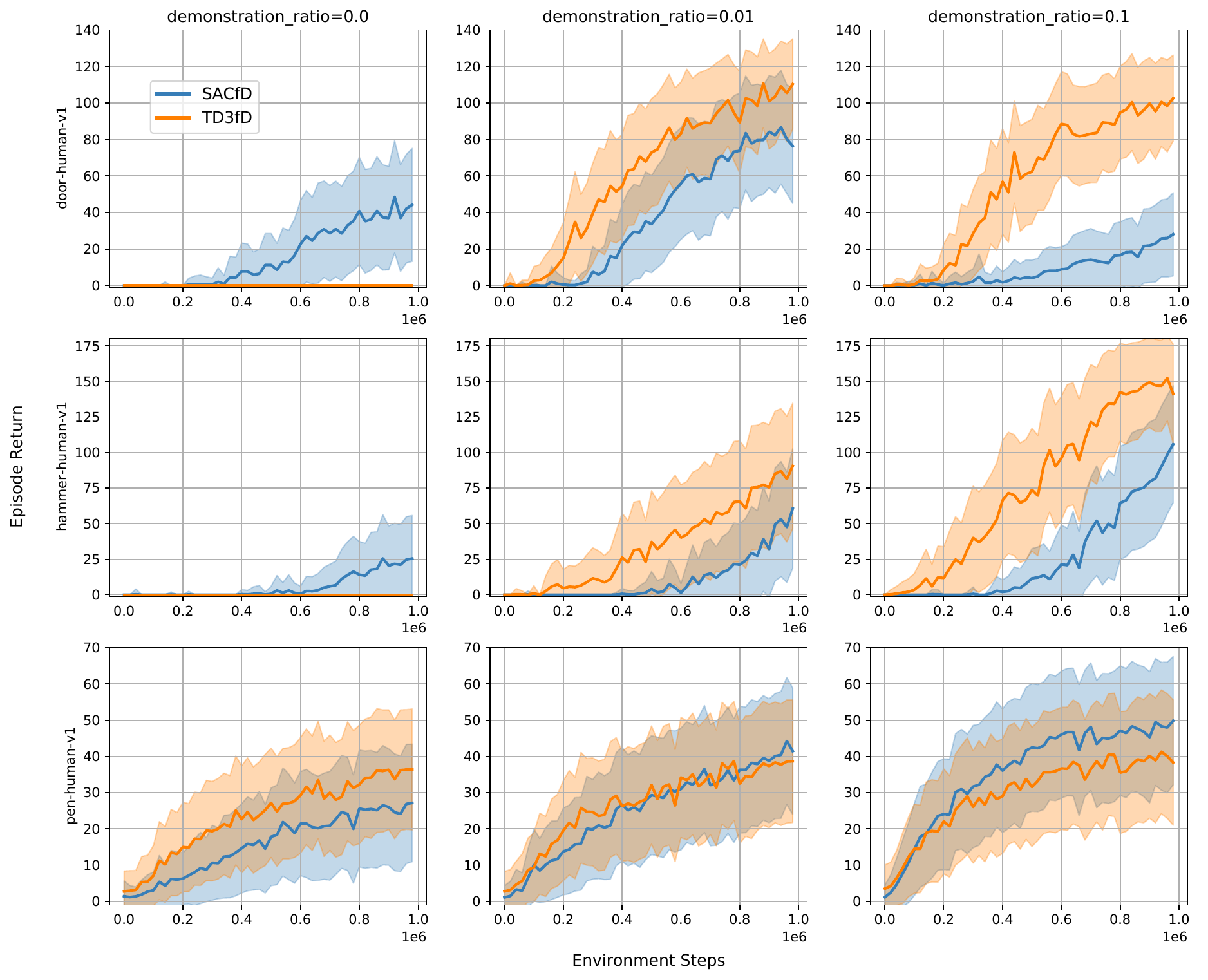}
    \caption{\textbf{Learning from Demonstrations.} Mean and standard deviation of the episode return obtained by the
    evaluation policies over 25 seeds in 1M environment steps. Each column
    corresponds to a different ratio of demonstration transitions in the
    replay buffer against agent's experienced transitions. Each row is a
    different environment.}
    \label{fig:xp:rlfd}
\end{figure}

\pagebreak
\paragraph{Learning from Demonstrations} We evaluate TD3fD and SACfD on a
sparse version of Adroit environments which rewards the agent with 1 if the
goal is achieved and 0 otherwise. We use the same demonstrations as
\citet{rajeswaran2017learning}, which consists in 25 trajectories acquired by
a human controller through a virtual reality system. In Figure
\ref{fig:xp:rlfd} we show the influence of the ratio of demonstration
transitions in the replay buffer against the agent's historical transitions.
The learning curves clearly illustrate that using demonstration data lead to
better performance. We did not include learning curves on the Relocate task
since none of the methods was able to get non-zero performance.

\begin{SCfigure}[][htb!]
  \caption{\textbf{Speed Comparison.} Mean and standard deviation of the
  episode return against the training walltime obtained by R2D2's evaluation
  policy on Pong over 2 seeds in 15M actor steps run with 256 actors. The
  experiment is repeated identically on three different machines: a GPU V100,
  a TPU 1x1 and a TPU 2x2.}
  \label{fig:speed}
  \includegraphics[width=0.4\textwidth]{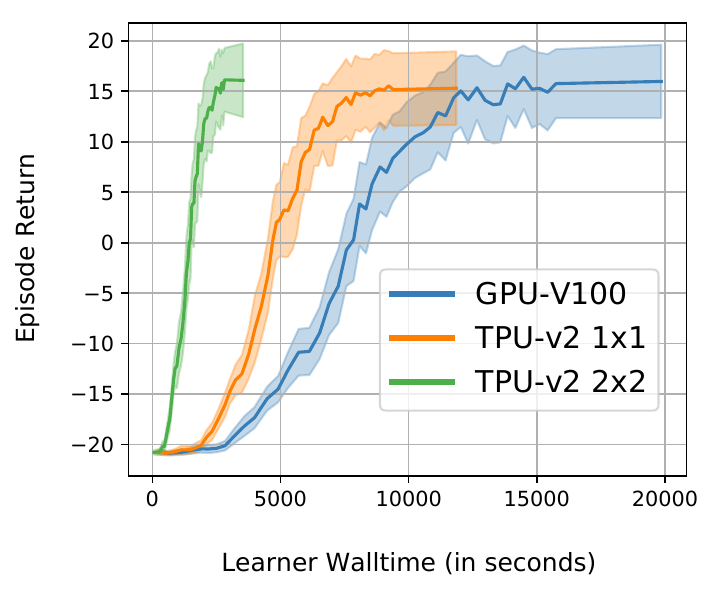}
\end{SCfigure}

\paragraph{Speed}
In order to showcase how easily Acme agents can be sped up by simply scaling
the learner hardware, we repeated an experiment three times on three different
machines: a GPU-V100, a TPU-v2 1x1 and a TPU-v2 2x2. To be clear, the agent
and its algorithmic logic is identical in all three cases, only the hardware
its learner is running on varies. For this comparison we focussed on Pong. In
order to ensure that the experiment is bottlenecked by the learner's SGD step,
we launched distributed R2D2 experiments with 256 actors and scaled the ResNet
torso up to have channels of sizes $(64, 128, 128, 64)$ per group instead of
$(16, 32, 32)$. The parameters are otherwise the default ones presented in
Figure \ref{fig:xp:drl}. As expected, the three experiments lead to the same
episode return, yet TPU-v2 1x1 yields a \textbf{1.7x speed-up} over GPU-V100,
while running on the full 2x2 topology leads to a \textbf{5.6x speed-up} over
GPU. Results are presented on Figure~\ref{fig:speed}

\section{Related work}
\label{sec:related-work}

Numerous open-source software libraries and frameworks have been developed in
recent years. In this section we give a brief review of recent examples, and
situate Acme within the broader context of similar projects. OpenAI baselines
\citep{dhariwal2017openaibaselines} and TF-Agents
\citep{guadarrama2018tfagents} are both examples of established deep RL
frameworks written in TensorFlow 1.X. They both strive to express numerous
algorithms in single-process format. Dopamine \citep{castro2018dopamine} is a
framework focusing on single-process agents in the DQN \citep{mnih2015atari}
family, and various distributional variants including Rainbow
\citep{hessel2018rainbow}, and Implicit Quantile Networks
\citep{dabney2018iqn}. Fiber \citep{zhi2020fiber} and Ray
\citep{moritz2017ray} are both generic tools for expressing distributed
computations, similar to Launchpad, described below. ReAgent
\citep{gauci2018horizon} is primarily aimed at offline/batch RL from large
datasets in production settings. SEED RL \citep{espeholt2019seed} is a highly
scalable implementation of IMPALA \citep{espeholt2018impala} that uses batched
inference on accelerators to maximize compute efficiency and throughput.
Similarly, TorchBeast \citep{kuttler2019torchbeast} is another IMPALA
implementation written in Torch. SURREAL \citep{fan2018surreal} expresses
continuous control agents in a distributed training framework. Arena
\citep{song2019arena} is targeted at expressing multi-agent reinforcement
learning.

The design philosophy behind Acme is to strike a balance between simplicity
and that of modularity and scale. This is often a difficult target to
hit---often it is much easier to lean heavily into one and neglect the other.
Instead, in Acme we have designed a framework and collection of agents that
can be easily modified and experimented with at small scales, or expanded to
high levels of data throughput at the other end of the spectrum.

\section{Conclusion}

This work introduces Acme, a modular light-weight framework that supports
scalable and fast iteration of research ideas in RL. Acme naturally supports
both single-actor and distributed training paradigms and provides a variety of
agent baselines with state-of-the-art performance. This work also represents a
second version of the paper which has seen an increase focus on modularity as
well as additional emphasis on offline and imitation algorithms. The number of
core algorithms implemented as part of Acme has also increased.

By providing these tools, we hope that Acme will help improve the status of
reproducibility in RL, and empower the academic research community with simple
building blocks to create new RL agents. Additionally, our baselines should
provide additional yardsticks to measure progress in the field. We are excited
to share Acme with the research community and look forward to contributions
from everyone, as well as feedback to keep improving and extending Acme.

\section{Author Contributions \& Acknowledgments}

Matt Hoffman, Gabriel Barth-Maron, John Aslanides, and Bobak Shahriari
co-designed and implemented the first version of Acme that was released on
June 1, 2020. This was based heavily on an initial prototype designed by
Gabriel Barth-Maron, who provided much of the implementation for this initial
prototype, and Matt Hoffman. Matt and Gabriel also contributed heavily to the
design of Launchpad, which enabled distributed agents. Gabriel was also an
important contributor to open-sourcing Reverb, which is used for experience
replay. John Aslanides played a significant role in the underlying design for
Actors, policies, and environment loops; he also contributed both the first
DQN implementation as well as designing the first JAX agents in Acme (which at
that point were purely TensorFlow). Matt and Bobak later updated these designs
to use JAX more centrally, helped integrate the Builder mechanism, and
contributed ongoing maintenance and design. Matt was also the primary author
of the first and second versions of the accompanying paper and Bobak
coordinated these experiments. Bobak contributed the MPO agent for both
TensorFlow and JAX and was heavily involved in core agent design.

Nikola Momchev contributed the Builder design along with its first
implementation with contributions from Danila Sinopalnikov and Piotr Stanczyk.
They worked to integrate this into Acme along with modular agents built using
this system. Their work also enabled the incorporation of a number of
imitation learning agents. They were also heavily involved in the continuing
design and ongoing maintenance of Acme. Danila Sinopalnikov, in addition,
contributed significantly to Acme’s agent testing infrastructure. Piotr
Stanczyk contributed to the joint experiment configuration between
single-process and distributed agents; he also contributed to ongoing
maintenance for both Launchpad and Reverb, including their integration with
Acme, and was instrumental in open sourcing Launchpad.

Sabela Ramos incorporated policy construction into the Builder interface as
well as implementing structured Adders. Anton Raichuk provided numerous speed
and efficiency improvements, implemented the first version of configurable
policy and evaluator callbacks, and implemented SAC.

Léonard Hussenot and Robert Dadashi contributed to a number of
agents---including TD3, PWIL, and SACfD---and were heavily involved in
designing the experimental setup for the second version of the paper. Léonard
led the coordination of experiments for the paper as well as writing the
experimental section and providing plots. Both Robert and Léonard were also
heavily involved in writing the paper as a whole, particularly the
experimental and algorithm-specific descriptions (Sections 4 and 5).

Gabriel Dulac-Arnold contributed to running continuous control experiments,
and helped with writing an early draft of Section 4, provided editing support,
as well as contributing important design feedback for future generalizations.
Manu Orsini, Alexis Jacq, Damien Vincent, Johan Ferret, Nino Vieillard
implemented or made improvements upon a number of agents including GAIL, DQN,
R2D2. Seyed Kamyar Seyed Ghasemipour made a number of agent improvements for
supporting large-scale PPO, alongside Acme infrastructure contributions for
supporting multi-learner agents. Sertan Girgin contributed bug fixes and a
number of helpers for experiment launchers. Olivier Pietquin provided support
and coordination throughout the second version of both the code and paper.

Feryal Behbahani helped document the first version of Acme, including writing
tutorial notebooks, and wrote algorithm descriptions in the accompanying
paper. Tamara Norman provided a number of efficiency improvements throughout
the initial codebase. Abbas Abdolmaleki tuned all MPO variants, as well as
being instrumental in its development, for both versions of the paper. Albin
Cassirer helped integrate Acme with Reverb and provided the first version of
Adders. Fan Yang helped integrate Acme with Launchpad. Kate Baumli provided
initial documentation, notebook tutorials, and the initial version of the JAX
R2D2 agent. Sarah Henderson provided support and coordination for releasing
the code and writing the paper. Abe Friesen contributed code and tuning for
the updated MPO agent. Ruba Haroun provided documentation and tutorials for
the second version. Alex Novikov provided a number of bug fixes and initial
offline agents for the first version of the paper. Sergio Gómez Colmenarejo
provided bug fixes and helped with snapshot and checkpoint code for the first
version. Serkan Cabi provided feedback and bug fixes for the initial JAX
version of D4PG. Caglar Gulcehre and Tom Le Paine provided offline algorithms
and experiments for both versions of the paper; Srivatsan Srinivasan
contributed to these algorithms for the second version. Andrew Cowie provided
bug fixes and an initial implementation of BCQ for the first version of the
paper. Ziyu Wang and Alex Novikov contributed to an initial version of CRR.
Bilal Piot provided feedback and design for the initial versions of R2D2 and
MPO. Nando de Freitas provided support and feedback for both versions of the
code and paper.

We would also like to thank Matthieu Geist and Adrien Ali Ta\"iga for their
valuable feedback on this manuscript. Thanks also to Yannis Assael and
Vaggelis Margaritis for a great deal of graphical feedback on this work and
Jackie Kay, David Budden, and Siqi Liu for help on an earlier version of this
codebase.

\clearpage
\bibliographystyle{apalike}
\bibliography{acme}

\clearpage
\appendix

\section{Algorithm Tag Descriptions}
\label{sec:tag-lookup}
\begin{itemize}
    \item \tagdiscrete \tagcontinuous describe the nature of actions space
    supported by the agents.
    \item \tagonpolicy \tagoffpolicy \tagoffline describe the training regimes
    of the agents, whether the learned policy is being played to gather data
    (on-policy), or another policy is being played to gather data (off-policy)
    or no data is being gathered (offline).
    \item \tagQnetwork \tagVnetwork \tagPnetwork describe the functions
    approximated with neural networks required by the agents.
    \item  \tagmc \tagbootstrapping describe the learning logic of the value,
    either by unrolling full trajectories (Monte-Carlo rollouts), or by using
    its own estimate to define the target (bootstrapping).
\end{itemize}
\end{document}